\documentclass{article} 
\usepackage{iclr2027_conference,times}

\usepackage{amsmath,amsfonts,bm}

\def\eqref#1{equation~\ref{#1}}

\def\1{\bm{1}}

\DeclareMathAlphabet{\mathsfit}{\encodingdefault}{\sfdefault}{m}{sl}
\SetMathAlphabet{\mathsfit}{bold}{\encodingdefault}{\sfdefault}{bx}{n}

\usepackage{hyperref}
\usepackage{url}

\usepackage{booktabs}
\usepackage{graphicx}
\usepackage{hyperref}
\usepackage{xurl}

\usepackage{lipsum}
\usepackage{colortbl}
\usepackage{wrapfig}
\usepackage{xspace}
\usepackage{multirow}
\usepackage{enumitem}

\usepackage{todonotes}

\usepackage{graphicx}
\usepackage{subcaption}

\title{Don't Stop Long-Context Continual Pre-Training Too Early: An Empirical Scaling Law}

\author{
Ziqing Qiao$^{1}$\thanks{Equal contribution.},
Xiaoyue Xu$^{2}$\footnotemark[1],
Sikui Zhang$^{2}$,
Yinuo Xu$^{1}$
Bingxiang He$^{1}$,
Chaojun Xiao$^{1}$\thanks{Corresponding authors.},
Xu Han$^{1}$\footnotemark[2]
\\
$^{1}$Department of Computer Science and Technology,
Tsinghua University, Beijing, China
\\
$^{2}$OpenBMB
\\
\texttt{qzq24@mails.tsinghua.edu.cn},
\texttt{\{xcj,han-xu\}@tsinghua.edu.cn}
}
\iclrfinalcopy
\begin{document}

\maketitle

\begin{abstract}
Long-context continual pre-training (LCCP) is widely used to improve long-context capability, but how its benefits scale with the number of tokens used for LCCP remains unclear.
We investigate this question through modeling, monitoring, and validating the benefits of scaling LCCP.
First, we establish an empirical additive power law relating long-context performance to model size and the respective token budgets of short-context pre-training and LCCP.
Experiments across five model scales and more than one hundred short- and long-context budget configurations show that the fitted law reproduces the observed trends and extrapolates well to held-out model scales.
Second, we introduce \textsc{Constant-Last4}, a training and observation protocol that averages the four most recent consecutive checkpoints from a constant-learning-rate run to enable stable monitoring along a single extensible training trajectory.
Under this protocol, the same scaling-law form fits well and accurately predicts the LCCP performance trajectory of a 1.22B-parameter model trained on approximately 100B LCCP tokens.
Finally, downstream evaluations along this trajectory show continued improvements in long-context performance, with no clear sign of saturation within the evaluated range.
These gains persist after both 128K context extension and supervised fine-tuning, with benefits extending to long-context understanding and agentic tasks.
Together, these results argue for continuing to scale LCCP rather than stopping early: its gains are stably observable along an extensible trajectory, quantitatively predictable, and transferable to downstream tasks.~\footnote{The code is released at \url{https://github.com/thunlp/dont-stop-lccp}.}
\end{abstract}
\section{Introduction}
Long-context capability is fundamental to language models, supporting complex long-horizon tasks and agentic workflows~\citep{team2026kimi, zeng2026glm, liu2025deepseek}. However, training on long sequences from the beginning of pre-training incurs substantial computational costs~\citep{grattafiori2024llama}. Long-context continual pre-training (LCCP) is therefore widely adopted, often as a brief stage following short-context pre-training. It uses longer sequences and a long-context-oriented data mixture to extend the context window and improve long-context performance~\citep{xiong2024effective,gao2025train}. A range of studies and models adopt relatively limited additional training to extend context windows~\citep{liu2024deepseek,fu2024data, hu2025longrecipe}.


As applications increasingly demand long-context understanding and sustained agentic interaction, some training recipes devote substantial token budgets to long-context pre-training~\citep{grattafiori2024llama,zeng2026glm}. Yet how long-context performance scales with the LCCP training budget remains insufficiently characterized. We therefore ask: \emph{Does continued LCCP yield sustained improvements in long-context capability, and can these gains be quantitatively predicted?}

\begin{figure}
    \centering
    \includegraphics[width=1\linewidth]{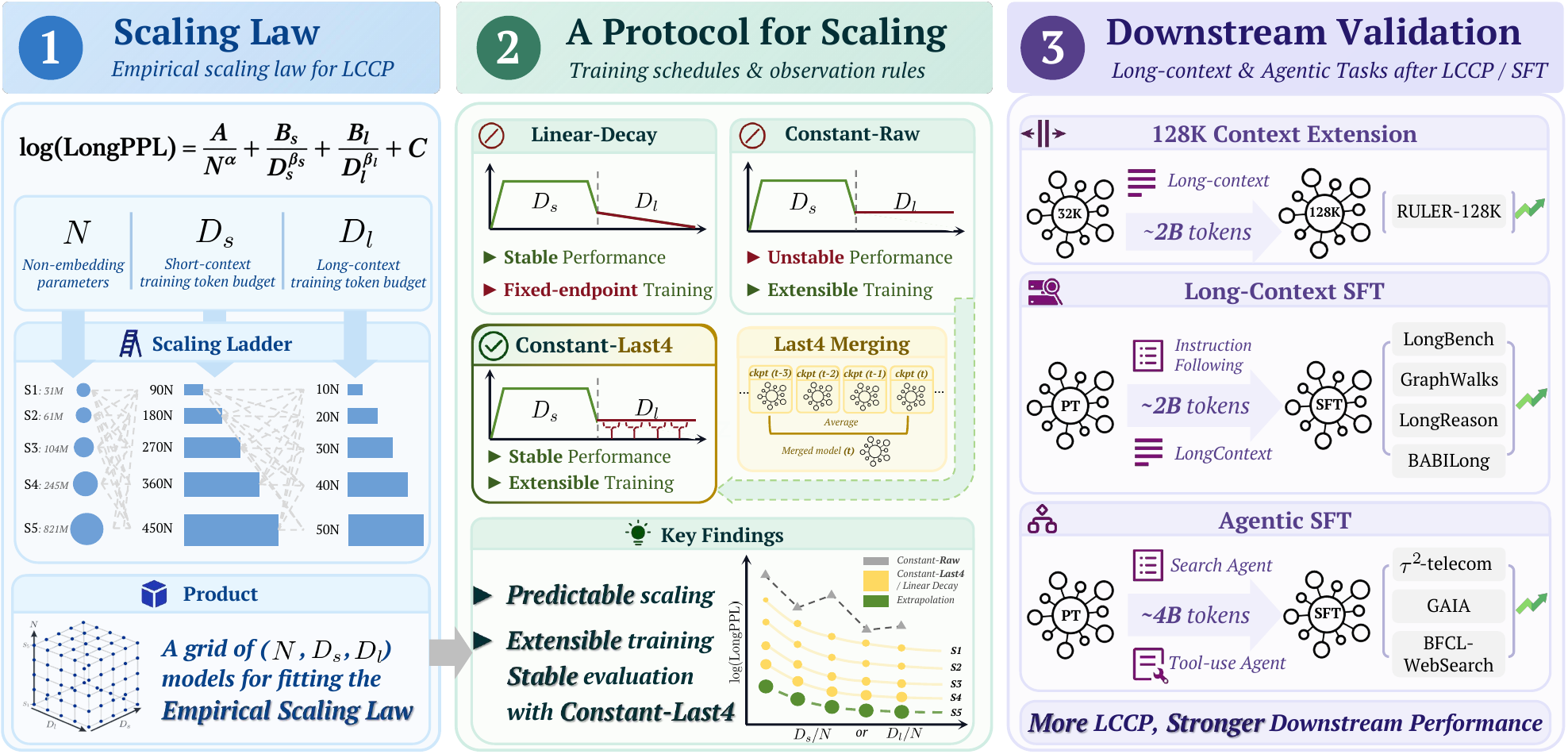}
    \caption{Overview of our framework for modeling, monitoring, and validating long-context performance during LCCP scaling. We develop an empirical scaling law for long-context performance, introduce \textsc{Constant-Last4} as an extensible training and observation protocol, and evaluate downstream performance gains after context extension and supervised fine-tuning.}
    \label{fig:overview}
\end{figure}

In this work, we address this question by \textbf{modeling, monitoring, and validating} the benefits of scaling up LCCP.
As illustrated in Figure~\ref{fig:overview}, we conduct a systematic empirical study of how long-context performance evolves with increasing LCCP budgets, spanning short-context pre-training, 32K LCCP, subsequent 128K context extension, and supervised fine-tuning(SFT).
We first establish an empirical scaling law for \textbf{modeling} long-context performance during LCCP, jointly accounting for model size, short-context pre-training tokens, and long-context training tokens.
We then introduce a stable observation protocol for extensible LCCP, enabling performance \textbf{monitoring} along a constant-learning-rate trajectory and, together with the scaling law, prediction from early checkpoints.
Finally, we conduct downstream evaluations on long-context and agentic tasks, thereby \textbf{validating} whether gains from additional LCCP translate into downstream improvements and persist through subsequent training stages.


\textbf{Modeling LCCP Scaling.}
Unlike pre-training from scratch, LCCP starts from a model that has already acquired basic capabilities, meaning that its long-context performance may depend on the preceding short-context pre-training stage~\citep{wu2025retrieval}. We therefore jointly model the non-embedding parameter count \(N\), short-context pre-training tokens \(D_s\), and long-context training tokens \(D_l\). Using LongPPL~\citep{fang2025wrong} as a proxy for long-context performance, we evaluate 109 training configurations across five model scales under a linear-decay learning-rate schedule. We find that an additive power law with an offset accurately captures the observed performance trends and extrapolates to held-out model scales under the evaluated budget settings, providing a quantitative basis for predicting long-context performance under different training budgets.

\textbf{Monitoring LCCP Scaling.}
To study continued LCCP scaling, we need a training and observation protocol that enables consistent performance monitoring along a single extensible trajectory.
Although linear-decay schedules provide stable endpoint performance, they require a predefined training budget and are therefore inconvenient for open-ended training. In contrast, a constant learning rate avoids this endpoint dependence, allowing cumulative long-context training tokens $D_l$ to serve as a common coordinate for monitoring training progress.


Raw constant-learning-rate checkpoints, however, exhibit substantial performance fluctuations and underperform budget-matched linear-decay models. Motivated by prior work on training dynamics and checkpoint merging~\citep{wen2024understanding,tian2026wsm}, we introduce the \textsc{Constant-Last4} protocol, which averages the four most recent checkpoints for evaluation. Across evaluated configurations, this protocol provides more stable observations and closely matches linear-decay performance. Under this protocol, the same scaling-law form fits well and extrapolates from smaller models to the approximately 100B-token LCCP trajectory of a 1.22B model. Moreover, when model size and the short-context training budget are fixed, the law simplifies to a power law in $D_l$ that predicts later LongPPL from a fit to early LCCP checkpoints. Together, \textsc{Constant-Last4} and the empirical scaling law support stable monitoring and performance prediction when scaling up LCCP.

\textbf{Validating Capability Gains.}
The practical value of the scaling law depends on whether the performance improvements it captures translate into gains in downstream long-context and agentic tasks.
We therefore validate the scaling trend along the 32K LCCP trajectory using RULER~\citep{hsieh2024ruler}, where aggregate performance improves as the long-context training budget increases, with no clear sign of saturation within the evaluated range.
Since LCCP is typically not the final training stage, we further evaluate checkpoints obtained at different LCCP budgets after 128K context extension and SFT for long-context and agentic tasks, using identical training procedures within each setting.
The benefits of LCCP scaling remain observable after both subsequent training stages and across multiple long-context and agentic benchmarks.
Together, these findings provide a quantitative basis for assessing the downstream benefits of continued LCCP.
\section{Related Work}
\label{sec:related}

\subsection{Scaling Laws for Large Language Models}
Scaling laws relate pre-training loss to model size and training data and characterize their trade-offs under fixed compute budgets~\citep{kaplan2020scaling,hoffmann2022training}.
Subsequent work has extended scaling-law analysis to continual pre-training, examining the effects of model size, additional training data, and data-mixture ratios~\citep{que2024d,gu2024cmr}, and characterizing learning dynamics~\citep{wang2025learning}, as well as to downstream performance prediction~\citep{ivgi2022scaling,hu2024predicting,chen2024scaling}.
However, scaling-law analyses of long-context performance remain comparatively limited.
Prior work has studied long-context scaling in hybrid architectures under single-stage pre-training~\citep{qiao2026rethinking}.
We instead study LCCP, jointly modeling long-context performance using model size and separate short- and long-context training budgets.

\subsection{Long-Context Continual Pre-Training}
LCCP typically starts from a short-context pre-trained model and extends its context window through positional adjustments and continued training on longer sequences~\citep{xiong2024effective, chen2023extending}, using a long-context-oriented data mixture to improve performance~\citep{gao2025train, fu2024data}.
Some methods emphasize token-efficient context extension~\citep{wu2025longattn, peng2024yarn}, whereas some leading large language models use long-context training budgets on the order of a trillion tokens~\citep{grattafiori2024llama,liu2025deepseek}.
These differing budget choices highlight the importance of understanding the benefits of continued LCCP and assessing whether training is sufficient.

Assessing the benefits of continued LCCP requires appropriate evaluation metrics.
LongPPL ~\citep{fang2025wrong} uses a reference model to identify tokens sensitive to distant context, providing a probabilistic measure of long-context performance.
Recent work analyzes LCCP learning dynamics at the behavioral, probabilistic, and mechanistic levels and introduces a retrieval-head-based metric~\citep{liang2026revealing}.
The findings suggest that behavioral scores can plateau while probabilistic and mechanistic metrics may continue to capture changes during training.
Complementing these analyses of learning dynamics, we focus on stable observation and quantitative prediction during continued LCCP.
Drawing on studies of learning-rate dynamics in pre-training~\citep{wen2024understanding,hagele2024scaling} and checkpoint averaging~\citep{tian2026wsm,sanyal2023early}, we develop \textsc{Constant-Last4}, an extensible protocol for LCCP training and LongPPL evaluation, and examine whether capability gains from additional LCCP persist after subsequent training stages.
\section{An Empirical Scaling Law for LCCP}
\label{sec:scaling_law}
To quantify how long-context performance scales during LCCP, we develop an empirical scaling law that jointly accounts for model size and the token budgets of the two training stages.
We evaluate both its fit to observed configurations and its predictive accuracy beyond the fitting range, across model scales and training budgets.

\subsection{Experimental Setup and Evaluation Metric}

Long-context performance after LCCP may depend not only on the amount of long-context training, but also on the model's short-context pre-training history~\citep{wu2025retrieval}.
This motivates treating the two stages separately rather than representing them by a single total token budget.
Specifically, we use three predictors: the non-embedding parameter count $N$, the number of short-context pre-training tokens $D_s$, and the number of long-context training tokens $D_l$.

\begin{wraptable}{r}{0.55\textwidth}
    \vspace{-10pt}
    \centering
    \caption{Model configurations.}
    \label{tab:model_configs}
    \vspace{-10pt}
    \scriptsize
    \setlength{\tabcolsep}{2pt}
    \resizebox{\linewidth}{!}{%
    \begin{tabular}{lrrrrrrr}
        \toprule
        \multirow{2}{*}{Scale} & \multirow{2}{*}{Layers} & \multicolumn{2}{c}{Dimensions} & \multicolumn{2}{c}{Heads} & \multicolumn{2}{c}{Parameters} \\
        \cmidrule(lr){3-4} \cmidrule(lr){5-6} \cmidrule(lr){7-8}
        & & Hidden & FFN & Attn. & KV & Non-emb. & Total \\
        \midrule
        S1 & 12 & 512  & 1,280 & 8  & 2  & 31M  & 165M \\
        S2 & 15 & 640  & 1,600 & 10 & 2  & 61M  & 228M \\
        S3 & 18 & 768  & 1,920 & 12 & 2  & 104M & 305M \\
        S4 & 24 & 1,024 & 2,560 & 16 & 2  & 245M & 513M \\
        S5 & 36 & 1,536 & 3,840 & 24 & 2  & 821M & 1.22B \\
        \bottomrule
    \end{tabular}%
    }
    \vspace{-10pt}
\end{wraptable}
To study LCCP scaling, we train models at five scales, S1--S5, with $N$ ranging from 31M to 821M, as shown in Table~\ref{tab:model_configs}. 
As for the training budget, for S1--S4, we use 
$$
\begin{aligned}
D_s/N &\in \{90, 180, 270, 360, 450\} \\
D_l/N &\in \{10, 20, 30, 40, 50\}
\end{aligned}
$$ 
yielding 100 configurations.
For larger-scale extrapolation, we construct five S5 configurations along the
diagonal $D_s=9D_l$, with $(D_s/N,D_l/N)\in
\{(90,10),(180,20),(270,30),(360,40),(450,50)\}$, and four off-diagonal
configurations, $(D_s/N, D_l/N)\in\{(90,50), (180,40), (360,20), (450,10)\}$, to provide
broad coverage of the configuration grid.

For the training setup, the short-context stage uses a sequence length of 4K and the WSD learning-rate schedule~\citep{hu2024minicpm}. 
The ratio between the number of tokens in the warmup+stable and decay
phases is 8:1, and $D_s$ includes tokens from all WSD phases. 
The LCCP stage uses a sequence length of 32K and a long-context data mixture. 
Following practices reported for open-source models~\citep{olmo2025olmo, bakouch2025smollm3},  we use a linear-decay learning-rate schedule during LCCP to obtain stable model performance. 
The learning rate starts from the value at the end of the short-context stage and decays to one tenth of that value. 
More details of training are provided in Appendix~\ref{app:training}.          

For the fitting target, we use LongPPL~\citep{fang2025wrong} as a proxy for long-context performance. Following the original protocol, we use Llama-3.1-8B~\citep{grattafiori2024llama} as the reference model and set the local chunk length to 4K tokens. For each token $x_t$, let
$\ell_{\mathrm{ref}}^{\mathrm{full}}(x_t)$ and
$\ell_{\mathrm{ref}}^{\mathrm{chunk}}(x_t)$ denote the
negative log-likelihood under the full context and the local chunk,
respectively. We select tokens that are well predicted under the full context and depend substantially on distant context:
\[
\Delta_t =
\ell_{\mathrm{ref}}^{\mathrm{chunk}}(x_t)
-
\ell_{\mathrm{ref}}^{\mathrm{full}}(x_t),
\qquad
\mathcal{K}_{\mathrm{long}}
=
\left\{
t:
\ell_{\mathrm{ref}}^{\mathrm{full}}(x_t)<2,\;
\Delta_t>2
\right\}.
\]
For an evaluated model $M$, LongPPL is computed from its
full-context loss only on $\mathcal{K}_{\mathrm{long}}$:
\[
\mathrm{LongPPL}(M)
=
\exp\left(
\frac{1}{|\mathcal{K}_{\mathrm{long}}|}
\sum_{t\in\mathcal{K}_{\mathrm{long}}}
\ell_M^{\mathrm{full}}(x_t)
\right).
\]
We use the same $\mathcal{K}_{\mathrm{long}}$ set for all models.
The evaluation sequences contain 32K tokens and are constructed from GovReport~\citep{huang2021efficient}. We compute LongPPL for all 109
training configurations and use $\log(\mathrm{LongPPL})$ as the target
for subsequent scaling-law fitting, consistent with previous work~\citep{qiao2026rethinking}.
Further details of the evaluation data are provided in Appendix~\ref{app:ppl_metric}.

\subsection{An Additive Two-Stage Scaling Law}

\begin{figure}[t]
\centering
\includegraphics[width=0.6\textwidth]{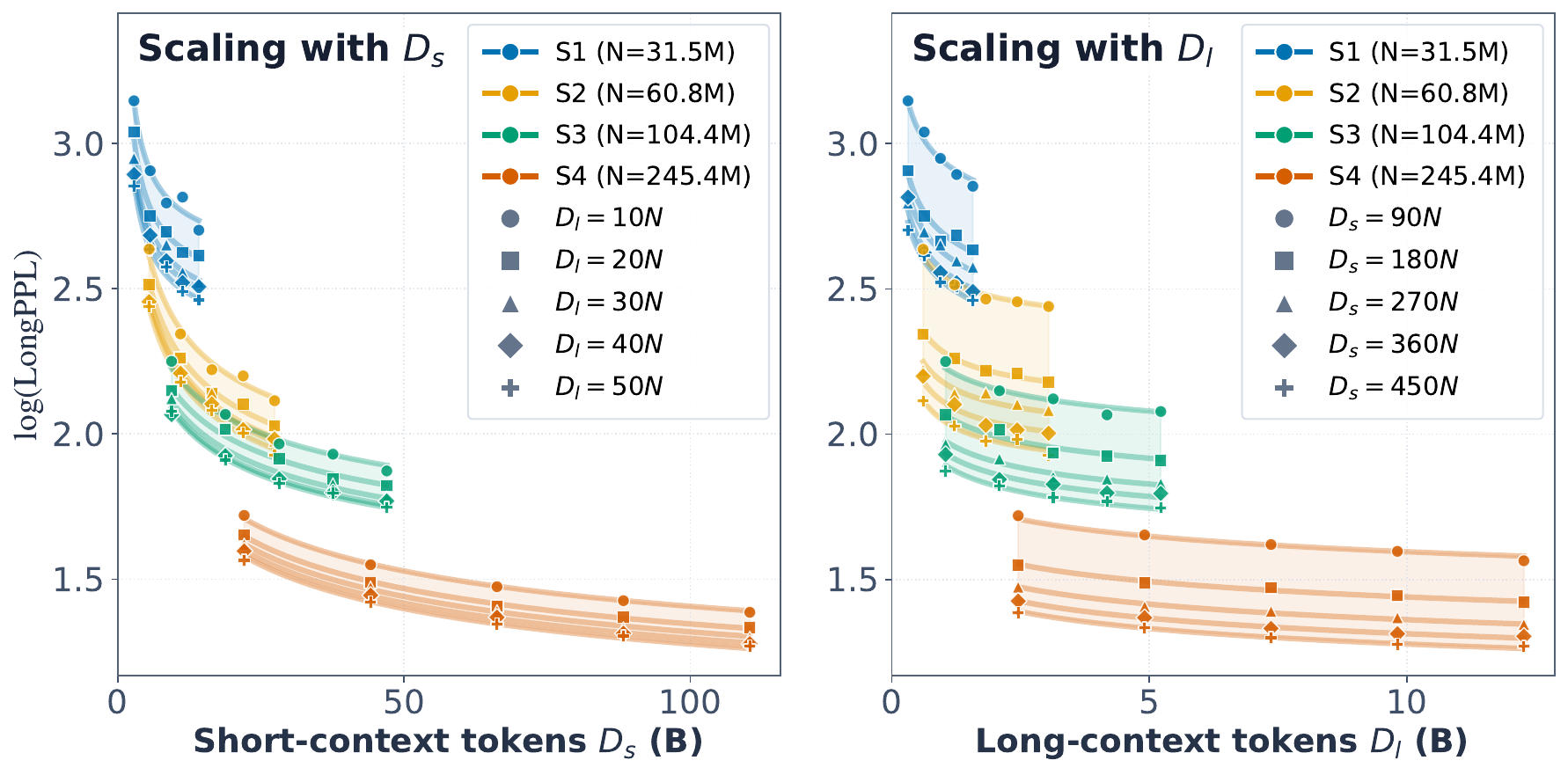}
\caption{$\log(\mathrm{LongPPL})$ versus $D_s$ and $D_l$, varying one budget while holding model size and the other budget fixed. Both trends are consistent with power-law decay with an offset.}
\label{fig:scaling_law_simple}
\vspace{-10pt}
\end{figure}

\textbf{Observation.}
To guide the choice of functional form, we examine how $\log(\mathrm{LongPPL})$ changes with $D_s$ or $D_l$ while holding the other two predictors fixed. 
As shown in Figure~\ref{fig:scaling_law_simple}, the observed trends are consistent
with power-law decay with an offset across the tested model scales and budgets. Guided by these trends and the additive structure of the Chinchilla scaling law~\citep{hoffmann2022training}, we consider the following formulation
\begin{equation}
    \log(\mathrm{LongPPL}) = \frac{A}{N^{\alpha}} + \frac{B_s}{D_s^{\beta_s}} + \frac{B_l}{D_l^{\beta_l}} + C,
    \label{eq:lccp_scaling_law}
\end{equation}
where $A$, $B_s$, and $B_l$ are scale coefficients, $\alpha$, $\beta_s$, and $\beta_l$ are scaling exponents, and $C$ is an offset. 
All seven parameters are fitted empirically.

\begin{figure}[t]
    \centering
    \includegraphics[width=0.95\linewidth]{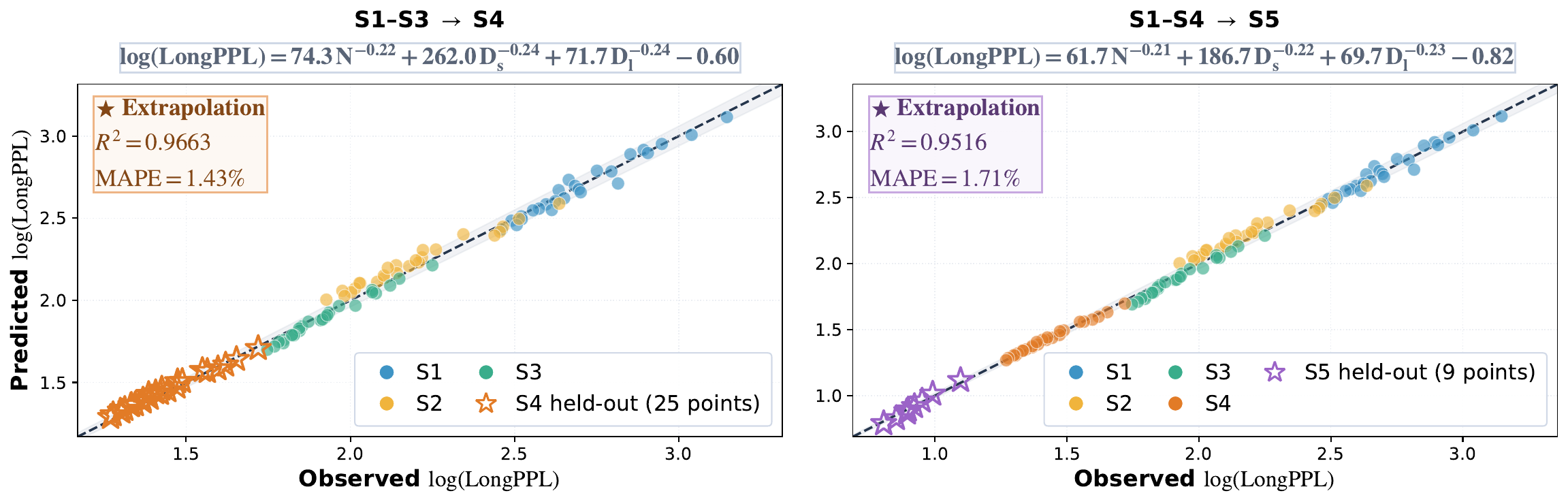}
    \caption{Scale-extrapolation parity plots for fits from S1--S3 to S4 (left) and S1--S4 to S5 (right), with the fitted laws shown above each panel. Dashed lines denote ideal predictions ($y=x$). $R^2$ and MAPE are computed in $\log(\mathrm{LongPPL})$ space on held-out configurations.}
    \label{fig:scaling_extrapolation_parity}
\end{figure}

\noindent\textbf{Fitting and scale extrapolation.}
We evaluate the fitting and scale-extrapolation performance of Eq.~\ref{eq:lccp_scaling_law} using two held-out-scale experiments.
First, we fit the scaling law using the 75 configurations from S1--S3 and evaluate it on the 25 configurations from S4. 
Second, we fit it using the 100 configurations from S1--S4 and evaluate it on the 9 configurations from S5.
In each experiment, we exclude configurations at the target model scale from fitting.
As shown in Figure~\ref{fig:scaling_extrapolation_parity}, Eq.~\ref{eq:lccp_scaling_law} fits the observed data well and extrapolates to both held-out model scales, achieving $R^2 > 0.95$ and mean absolute percentage error (MAPE) below $2\%$ in $\log(\mathrm{LongPPL})$ space. Precise fitted parameter values are reported in Appendix~\ref{app:fitting}.

Notably, the fitted offset satisfies $C<0$. Since $\log(\mathrm{LongPPL})$ is a mean negative log-likelihood and is nonnegative, we interpret Eq.~\ref{eq:lccp_scaling_law} as a finite-range empirical law rather than a valid asymptotic law. Nevertheless, our extrapolation tests extend substantially beyond the fitting range: S5 has more than three times the parameter count of S4, the largest model used for fitting, with token budgets scaling by the same factor at matched $D_s/N$ and $D_l/N$.

We also compare several alternative formulations, including variants with no offset or a nonnegative offset, formulations using effective data or the combined budget $D=D_s+D_l$, and formulations with interaction terms among the three predictors.
Among these formulations, Eq.~\ref{eq:lccp_scaling_law} attains the best overall extrapolation performance; details are provided in Appendix~\ref{app:formula}.
These comparisons provide additional empirical support for Eq.~\ref{eq:lccp_scaling_law} as a predictive model of long-context performance during LCCP within the evaluated range.
\section{An Extensible Training and Observation Protocol}

Under the linear-decay setup, we successfully fit an additive power-law model for $\log(\mathrm{LongPPL})$, which suggests that increasing the LCCP training budget can yield continued and predictable benefits within the studied range. 
Although the linear-decay schedule provides stable model performance, the learning-rate schedule is tied to a prespecified training endpoint, making it less convenient for open-ended LCCP scaling. 
In this section, we therefore explore a training and observation protocol that supports both extensible training and stable performance evaluation, facilitating the application of the scaling law to LCCP scaling in practice.

\subsection{Constant-Last4: Constant Learning Rate with Model Merging}
\label{sec:merge}
We use a constant learning rate to support open-ended LCCP without prespecifying a final training budget~\citep{liu2024deepseek,blakeman2026nemotron}.
The cumulative number of long-context training tokens $D_l$ then serves as a natural coordinate along a single training trajectory, which brings two practical benefits: additional long-context data can be incorporated by resuming from the latest checkpoint,
and model performance can be evaluated at any training budget without running a separate experiment for each budget.
To mitigate the elevated and fluctuating evaluation loss that can arise under constant-learning-rate training~\citep{wen2024understanding}, we consider checkpoint merging without additional training~\citep{tian2026wsm}.
We treat merging as an \emph{observation} mechanism: evaluation uses merged checkpoints, while the training trajectory remains unchanged.

\textbf{Observation rules and metrics.}
We take the raw constant-learning-rate checkpoint as the simplest observation rule and compare it with a family of checkpoint-merging methods.
As the reference for all rules, we use the \textsc{Linear-Decay} checkpoint trained with the same $(N, D_s, D_l)$.

For judging observation rules, we retain LongPPL as the long-context metric and introduce ShortPPL as a complementary diagnostic.
Reusing the same reference model and token-level statistics as in LongPPL, we select tokens for which the local chunk yields a lower loss than the full context and compute perplexity over these positions:
\[
\mathcal{K}_{\mathrm{short}}
=
\left\{
t:\Delta_t<0,\;
\ell_{\mathrm{ref}}^{\mathrm{full}}(x_t)<2
\right\},
\qquad
\mathrm{ShortPPL}(M)
=
\exp\left(
\frac{1}{|\mathcal{K}_{\mathrm{short}}|}
\sum_{t\in\mathcal{K}_{\mathrm{short}}}
\ell_M^{\mathrm{full}}(x_t)
\right).
\]
Since the selected tokens do not benefit from distant context, ShortPPL primarily reflects general language-modeling performance, and thus serves as a complementary metric.

\begin{figure}[t]
    \centering
    \begin{subfigure}[b]{0.40\linewidth}
        \centering
        \includegraphics[width=\linewidth]{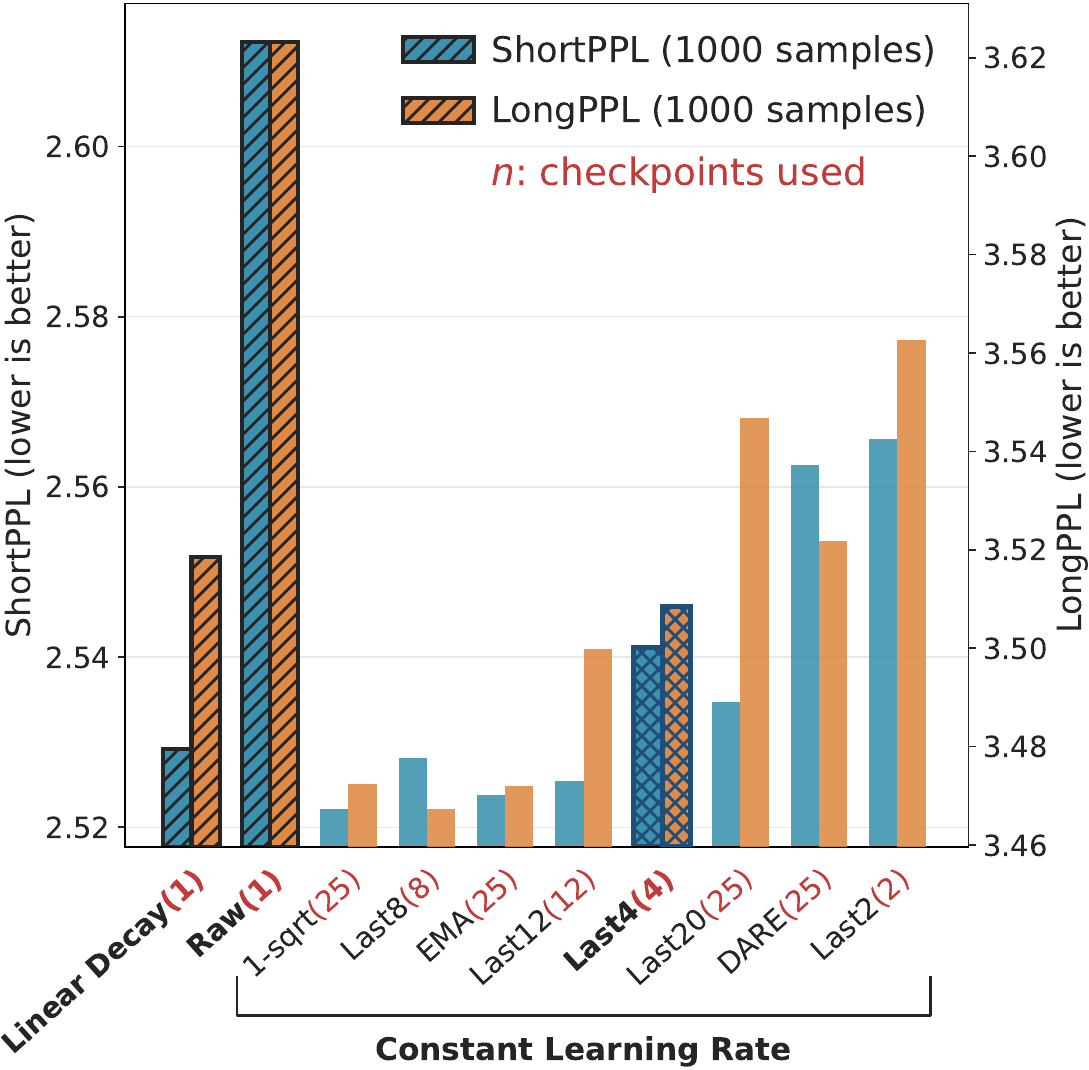}
        \caption{Observation rules}
        \label{fig:merge_methods}
    \end{subfigure}
    \hfill
    \begin{subfigure}[b]{0.59\linewidth}
        \centering
        \includegraphics[width=\linewidth]{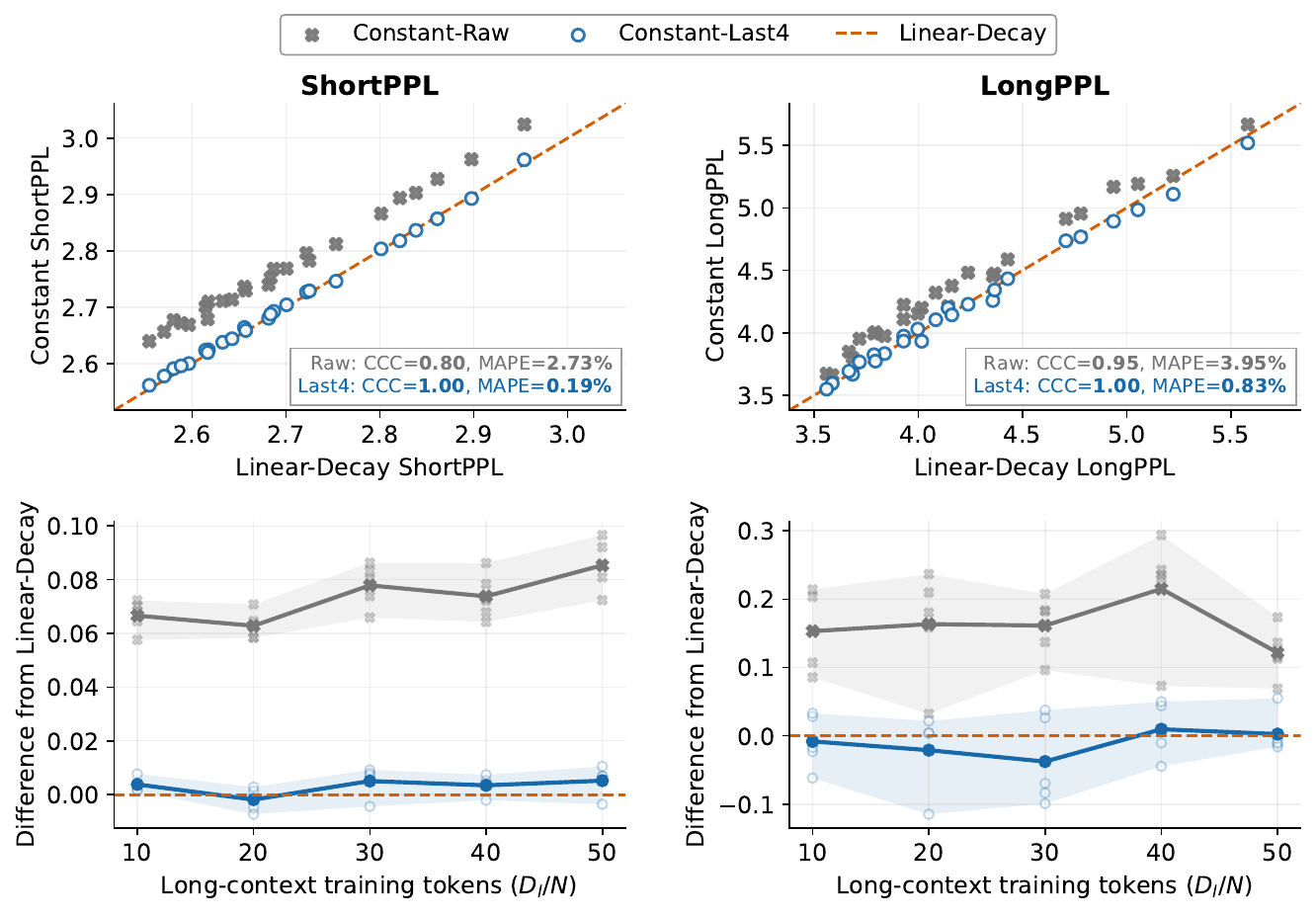}
        \caption{Training and observation protocols}
        \label{fig:lr_schedule}
    \end{subfigure}

    \caption{Comparison of training and observation protocols using ShortPPL and LongPPL. (a) Different observation rules on S4 with $D_s=450N$ and $D_l=50N$. (b) \textsc{Constant-Raw}, \textsc{Constant-Last4}, and matched \textsc{Linear-Decay} across 25 S4 configurations. \textsc{Constant-Raw} yields higher and noisier values for both metrics, whereas \textsc{Constant-Last4} closely matches \textsc{Linear-Decay}.
    CCC denotes Lin's concordance correlation coefficient~\citep{Lin1989ACC}.
    }
    \label{fig:checkpoint_merge}
\end{figure}

\textbf{Comparison setup.} Starting from the S4 short-context checkpoint with $D_s=450N$, we continue constant-learning-rate training to $D_l=50N$ and denote the resulting raw model by \textsc{Constant-Raw}.
Along the trajectory, we save checkpoints at intervals of $2N$ in $D_l$, yielding a pool of 25 checkpoints.
Within this pool we evaluate several representative merging methods: \textsc{Last}$M$, which assigns equal weight to the final $M$ consecutive checkpoints~\citep{sanyal2023early}, the recency-weighted profiles \textsc{1-Sqrt}~\citep{tian2026wsm} and \textsc{EMA}, and \textsc{DARE}~\citep{yu2024language}, a task-vector method. Additional implementation details are provided in Appendix~\ref{app:merging}.

\textbf{Observation rule selection.} As shown in Figure~\ref{fig:merge_methods}, \textsc{Constant-Raw} has substantially higher LongPPL and ShortPPL than the budget-matched \textsc{Linear-Decay} checkpoint, whereas checkpoint merging recovers most of this gap on both metrics.
Considering performance, simplicity, and storage requirements, we select \textsc{Last4} as the default observation rule: it averages only the four most recent consecutive checkpoints while preserving the latest raw checkpoint for continued training.

To further assess the stability of \textsc{Last4} merging, we compare \textsc{Constant-Raw} and \textsc{Constant-Last4} with \textsc{Linear-Decay} across all S4 configurations in Figure~\ref{fig:lr_schedule}.
Across training budgets, \textsc{Constant-Last4} closely matches the corresponding \textsc{Linear-Decay} checkpoint on both ShortPPL and LongPPL.
In summary, \textsc{Constant-Last4} combines the extensibility of constant-learning-rate training with the stable performance of a decayed endpoint in our setting, providing a practical training and observation protocol for LCCP scaling.

\subsection{Scaling-Law Validation under Constant-Last4}
\label{sec:scaling_last4}
For the new \textsc{Constant-Last4} protocol, we need to test whether the empirical scaling law Eq.~\ref{eq:lccp_scaling_law} remains valid under it. We adopt a $(D_s/N, D_l/N)$ budget grid for S1--S4 similar to the one in Section~\ref{sec:scaling_law}. The realized token counts and details are provided in Appendix~\ref{app:32k_lccp}.
Because a single constant-learning-rate trajectory provides checkpoints at multiple $D_l$ values, one run covers several long-context training budgets and reduces the experimental cost relative to \textsc{Linear-Decay}. 
We run each LCCP trajectory independently three times from the same short-context checkpoint and use the mean LongPPL across runs for S1--S4 models.

We follow the scale-extrapolation setup used in Section~\ref{sec:scaling_law}. 
First, we fit the scaling law on S1--S3 and evaluate it on S4. We then fit it on S1--S4 and extrapolate to the S5 trajectory. 
Whereas the \textsc{Linear-Decay} experiments used nine S5 configurations, here we extend pointwise scale extrapolation to prediction along an entire training trajectory.
Specifically, we fix the short-context budget at $D_s=450N$ and continue constant-learning-rate training to $D_l=120N$ (approximately 100B tokens). 
This setting provides dense observations of $\log(\mathrm{LongPPL})$ as a function of $D_l$ at a fixed model scale and short-context budget. For S5, we save checkpoints at $1N$ intervals and apply \textsc{Last4} to the four most recent checkpoints every $5N$, yielding 24 merged evaluation points. Appendix~\ref{app:merge-spacing} shows that the results are insensitive to checkpoint spacing within the evaluated range.

\begin{figure}[t]
    \centering
    \includegraphics[width=1\linewidth]{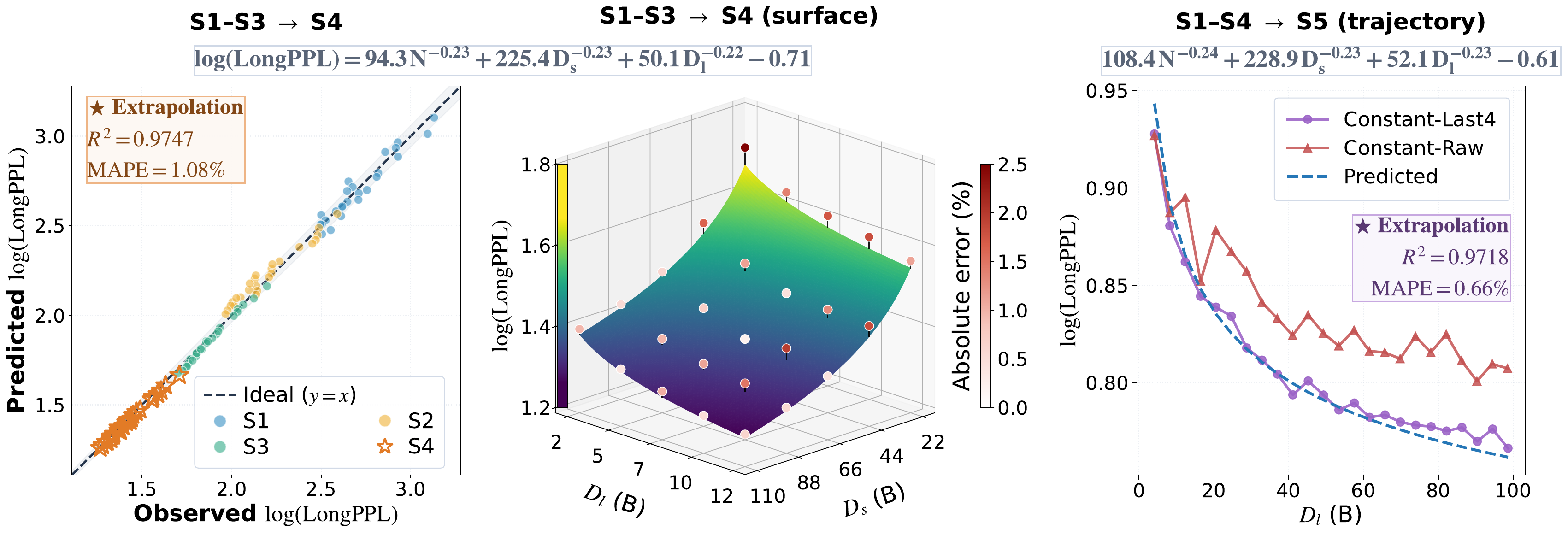}
    \caption{Scaling-law fitting and extrapolation under \textsc{Constant-Last4}: held-out S4 predictions from an S1--S3 fit (left and middle) and held-out S5 trajectory predictions from an S1--S4 fit (right). In the middle panel, dot colors indicate the absolute percentage error between predicted and measured $\log(\mathrm{LongPPL})$. Precise fitted parameter values are reported in Appendix~\ref{app:fitting}.}
    \label{fig:constant_scaling}
\end{figure}

As shown in Figure~\ref{fig:constant_scaling}, Eq.~\ref{eq:lccp_scaling_law} fits the \textsc{Constant-Last4} configurations well and extrapolates accurately to an unseen model scale. 
Notably, the S5 trajectory reaches approximately 100B long-context training tokens, more than eight times the largest absolute long-context budget in the fitting set, yet the scaling law accurately predicts $\log(\mathrm{LongPPL})$ across the entire trajectory.
Thus, the scaling law works under both \textsc{Linear-Decay} and \textsc{Constant-Last4} within the tested range.

\subsection{A Simplified Law for Fixed-Model LCCP}
\label{sec:simple_law}
\begin{wrapfigure}{r}{0.41\textwidth}
\vspace{-15pt}
\centering
\includegraphics[width=\linewidth]{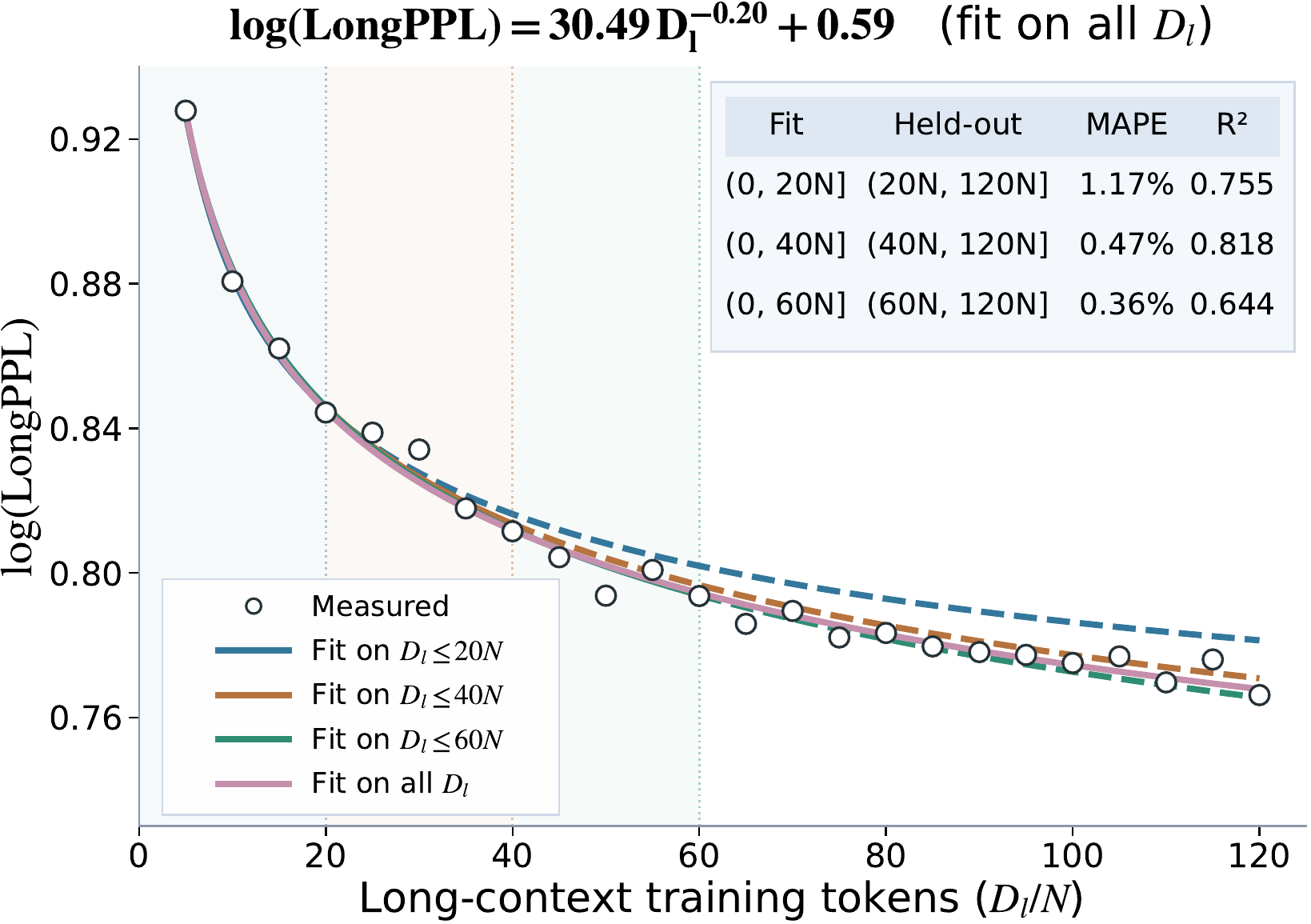}
\caption{Prefix fitting of $\log(\mathrm{LongPPL})$ along the S5 \textsc{Constant-Last4} run.}
\vspace{-20pt}
\label{fig:prefix}
\end{wrapfigure}
In practical LCCP, one often starts from a fixed model that has undergone substantial short-context pre-training and continues training it with additional long-context data.
In this setting, the main question is how performance evolves with $D_l$ and whether early
observations can predict the benefits of further training. When $N$ and $D_s$ are fixed, their contributions in Eq.~\ref{eq:lccp_scaling_law} can be absorbed into a single offset, yielding
\begin{equation}
    \log(\mathrm{LongPPL})
    = \frac{B_l}{D_l^{\beta_l}} + C',
    \label{eq:simple_law}
\end{equation}
where $C'=A/N^{\alpha}+B_s/D_s^{\beta_s}+C$.We re-estimate $B_l$, $\beta_l$ and $C'$ jointly from the given trajectory rather than taking them from the joint fit.
This lower-dimensional form is practical for fixed-model LCCP because it can be fitted using a small number of early checkpoints from a single training trajectory, avoiding the costly multi-scale, multi-budget training grid required for fitting the joint law.

We evaluate the S5 \textsc{Constant-Last4} trajectory with $D_s$ fixed at $450N$ and $D_l$ extending to $120N$, evaluating one merged checkpoint every $5N$.
We fit Eq.~\ref{eq:simple_law} using Huber loss on prefixes up to $20N$, $40N$, and $60N$, respectively, and evaluate each fit on all subsequent held-out checkpoints through $120N$.
As shown in Figure~\ref{fig:prefix}, even a short prefix provides a reasonable estimate of the remaining trajectory, suggesting that early observations can help estimate later LongPPL for a fixed-model LCCP trajectory. 
We further fit the same power-law form to ShortPPL and compare it with that of LongPPL to assess LongPPL's usefulness as a long-context proxy; details are provided in Appendix~\ref{app:longppl_analysis}.

\section{Downstream Validation of LCCP Scaling}
\label{sec:benchmark}
For both \textsc{Linear-Decay} and \textsc{Constant-Last4}, LongPPL decreases predictably as the LCCP budget increases. 
To test whether this improvement translates into long-context capability, we evaluate from three complementary perspectives: capability gains during 32K LCCP, their transfer to 128K context extension, and their preservation after SFT. 
We conduct these evaluations using checkpoints sampled from the S5 32K \textsc{Constant-Last4} trajectory introduced in Section~\ref{sec:scaling_last4}. Additional training and evaluation details for this section are provided in Appendix~\ref{app:benchmark}.

\subsection{Capability Gains during 32K LCCP}
\label{sec:ruler32k}
\begin{wrapfigure}{r}{0.4\linewidth}
    \vspace{-12pt}
    \centering
    \includegraphics[width=\linewidth]{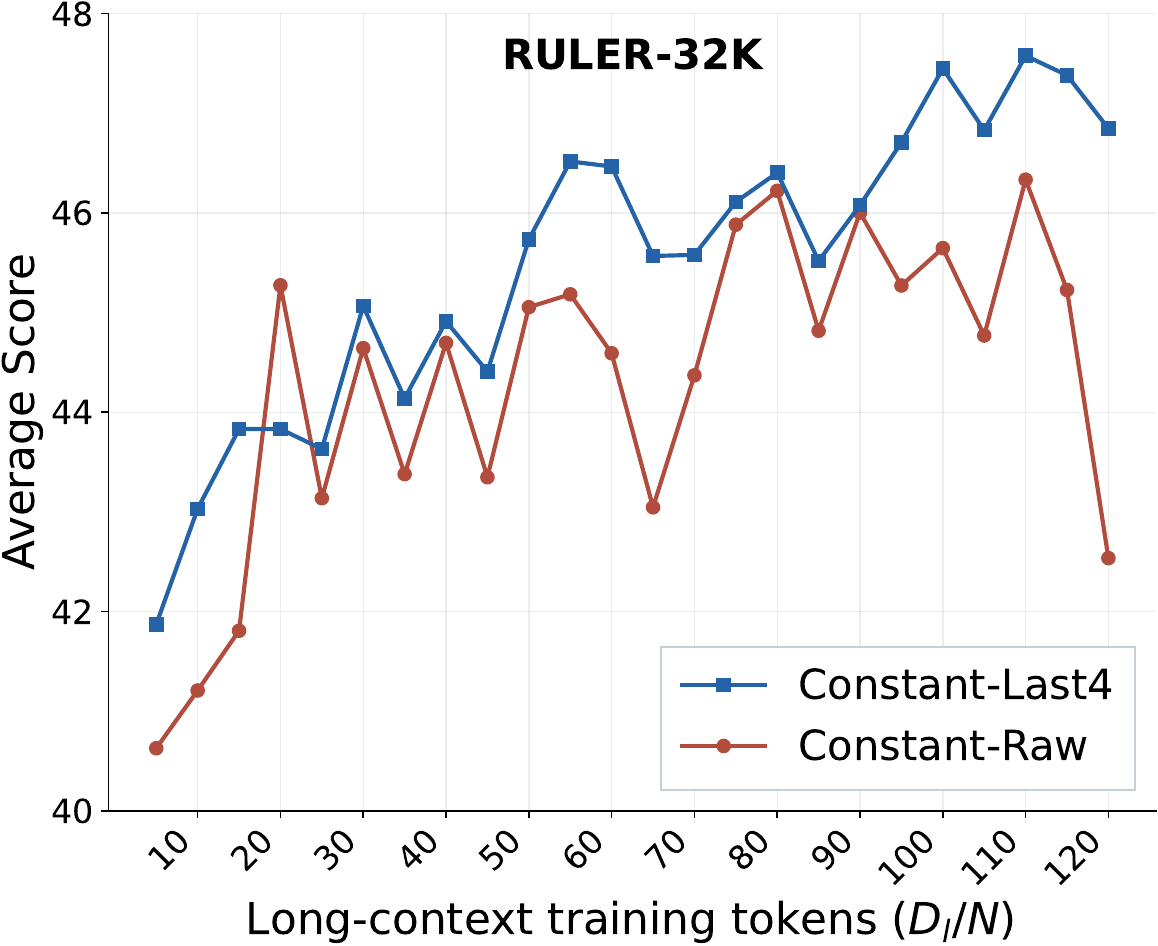}
    \caption{RULER-32K scores along the 32K LCCP trajectory of the S5 model.}
    \label{fig:ruler32k}
    \vspace{-15pt}
\end{wrapfigure}

To test whether the decrease in LongPPL translates into genuine long-context capability, we evaluate the LCCP trajectory of the S5 model using RULER~\citep{hsieh2024ruler}. We construct 500 samples for each subtask and evaluate both \textsc{Constant-Raw} and \textsc{Constant-Last4} checkpoints every $5N$ along the trajectory.

As shown in Figure~\ref{fig:ruler32k}, the RULER scores of \textsc{Constant-Last4} exhibit an overall upward trend throughout the observed LCCP trajectory, with no clear sign of saturation. \textsc{Constant-Raw} follows a similar trend but shows larger fluctuations and generally lower scores at comparable training budgets. 
Detailed per-subtask results in Appendix~\ref{app:ruler_32k} show that the aggregate gains are driven primarily by intermediate-difficulty retrieval subtasks. 
These results link continued LCCP scaling to improved long-context benchmark performance within the evaluated range, with \textsc{Constant-Last4} providing a more stable capability signal under a constant learning rate.

\subsection{Capability Gains after 128K Context Extension}
\label{sec:128k}

Multi-stage LCCP is common in practical training pipelines, where a model is first trained extensively at a moderate context length and then extended to a longer context with a smaller additional budget~\citep{grattafiori2024llama, mai_thinking_1}. 
To test whether gains from 32K LCCP transfer to a longer context, we select checkpoints at $D_l/N\in\{10,20,40,60,80,100,120\}$ from the previous S5 32K LCCP trajectory and apply the same 128K context-extension procedure to both \textsc{Constant-Raw} and \textsc{Constant-Last4} models. 
Specifically, we use ABF~\citep{xiong2024effective} to increase the RoPE base frequency from $500{,}000$ to $5{,}000{,}000$, use a 128K-stage data mixture with a higher proportion of long-context sequences than the 32K LCCP mixture, and continue training for 2.10B tokens at a sequence length of 128K. 
The learning rate starts from the constant rate used during 32K LCCP and follows a linear decay to one hundredth of its initial value.

\begin{figure}[t]
    \centering
    \includegraphics[width=1\linewidth]{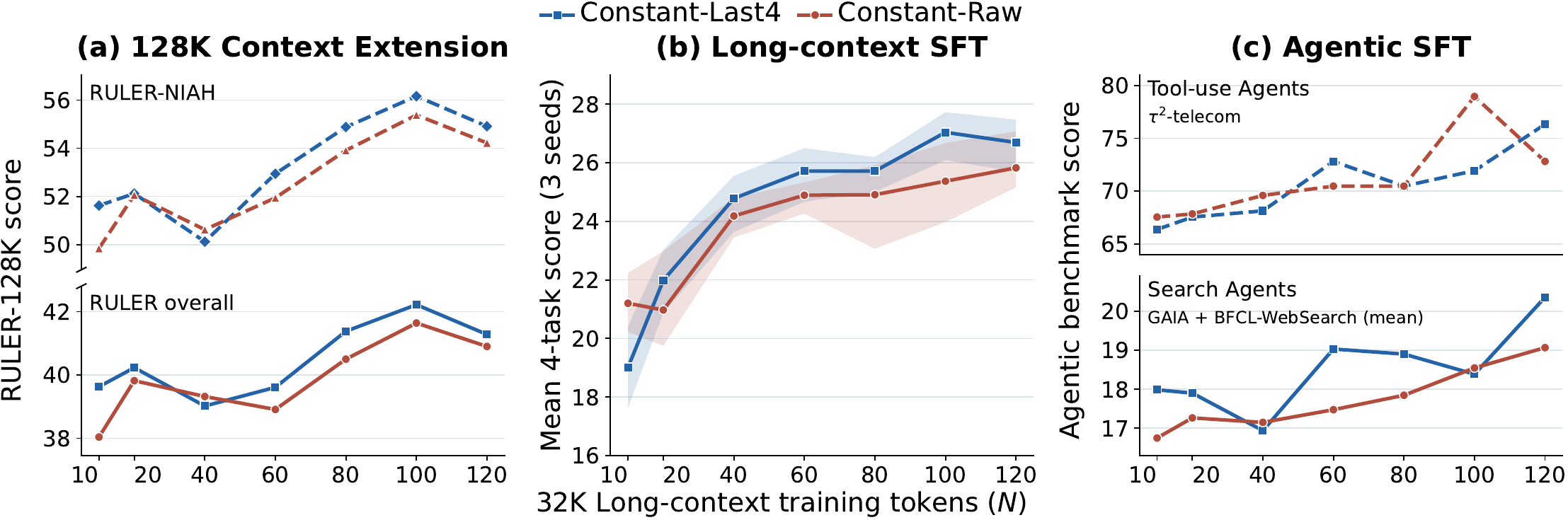}
    \caption{Downstream performance after 128K extension and SFT at varying 32K LCCP budgets. Shaded regions in (b) show the range across 3 training seeds. Across these subsequent training stages, larger 32K LCCP budgets generally lead to improved downstream performance.}
    \label{fig:downstream_transfer}
\end{figure}

For evaluation, we construct 100 128K-length samples per RULER subtask.
As shown in Figure~\ref{fig:downstream_transfer}(a), the performance of both \textsc{Constant-Raw} and \textsc{Constant-Last4} models generally improves with increasing 32K budget, especially on NIAH-related tasks. 
These results suggest that the benefits of a larger 32K LCCP budget carry over to the longer-context regime after context extension.

\subsection{Capability Gains after Supervised Fine-Tuning}
\label{sec:sft}

The goal of scaling LCCP is to improve downstream model performance. Since models typically undergo SFT to adapt their pretrained capabilities to practical use, we evaluate whether the benefits of larger LCCP budgets persist after SFT. We consider two settings, long-context SFT and agentic SFT, using the same checkpoints of the 32K LCCP trajectory of the S5 model as in Section~\ref{sec:128k}. Within each setting, we apply the same SFT procedure to checkpoints obtained at different LCCP budgets.
More training details are provided in Appendices~\ref{app:training}–~\ref{app:training-data}; evaluation protocols and supplementary results are provided in Appendix~\ref{app:sft}.

\noindent\textbf{Long-context SFT.}
We apply the same 2.10B-token SFT recipe to each checkpoint, using the non-thinking response format.
We evaluate the SFT models on four long-context benchmarks: LongBench~\citep{bai2024longbench}, GraphWalks~\citep{graphwalks}, LongReason~\citep{ling2025longreason}, and BABILong~\citep{kuratov2024babilong}, and report their equal-weight average.
As shown in Figure~\ref{fig:downstream_transfer}(b), downstream performance generally improves with the LCCP training budget under the same SFT procedure. Each point reports the mean over three independent SFT runs with different data seeds, with the shaded region indicating the minimum-to-maximum range across the three runs.

\noindent\textbf{Agentic SFT.}
We further examine whether the benefits extend to agentic scenarios. We consider two task categories: search agents, evaluated on GAIA~\citep{mialon2024gaia} and BFCL-WebSearch~\citep{patil2025bfcl}, and tool-use agents, evaluated on $\tau^2$-telecom~\citep{barres2025tau2benchevaluatingconversationalagents}. Each checkpoint undergoes the same 4.20B-token agentic SFT recipe with thinking mode enabled during training. Each benchmark score is averaged over three evaluation runs, and the two search benchmark scores are further averaged with equal weights.
As shown in Figure~\ref{fig:downstream_transfer}(c), performance in both categories generally improves with larger LCCP budgets, despite fluctuations across checkpoints.

Taken together, these evaluations show that, for the 1B-scale S5 model, the benefits of scaling 32K LCCP to approximately 100B tokens remain evident after 128K context extension and after SFT for both long-context and agentic tasks. These results support further LCCP scaling as a promising direction for improving downstream capabilities when sufficient long-context data are available.



\section{Conclusion}


We establish an empirical scaling law for LCCP that relates long-context performance to model size and separate token budgets for short-context pre-training and LCCP, while supporting extrapolation to held-out model scales and token budgets. We introduce \textsc{Constant-Last4}, enabling stable observation along an extensible trajectory. In our 1B-scale experiments, we observe continued improvements in long-context performance along the evaluated 32K LCCP trajectory, up to $\approx 100B$ tokens. These gains remain after 128K context extension and SFT, including downstream long-context and agentic tasks. Taken together, these findings establish continued LCCP as a predictable and practically effective path toward improving long-context and agentic capabilities.

\section*{AI use statement}

We used generative AI tools to assist with translation, correct grammatical and stylistic issues, improve the clarity and academic presentation of the manuscript, refine scientific visualizations, and format LaTeX tables and figures.
We did not use generative AI to generate data, formulate scientific claims or hypotheses, design experiments or develop methods, conduct a literature review or perform qualitative analysis, or generate citations. All AI-assisted work was reviewed and verified by the authors. The authors independently determined all interpretations and conclusions based on the original experimental results and take full responsibility for the final content of this work.

\section*{Ethics Statement}
This work studies the scaling behavior of long-context continual pre-training. Improved long-context and agentic capabilities may amplify existing language-model risks, including harmful outputs,
bias, privacy leakage, and misuse of automated tools. The use of large-scale training corpora also warrants attention to data provenance, licensing, and potentially sensitive content.
Our capability evaluations do not establish deployment safety; downstream applications require task-specific safety evaluation and appropriate safeguards. Further scaling should weigh capability gains against these risks and the computational and environmental costs of training.

\section*{Reproducibility Statement}
Sections~\ref{sec:scaling_law}--\ref{sec:benchmark}
describe our experimental design, scaling-law formulation,
and evaluation protocols. Appendix~\ref{app:setup}
documents metric computation and evaluation-set construction,
model and training configurations, realized token budgets,
hardware, and training-data composition and processing.
It also specifies the scaling-law fitting procedure,
held-out evaluation splits, and fitted parameters.
Appendix~\ref{app:merging} details the checkpoint-merging
methods and their comparison protocol.
Appendix~\ref{app:benchmark} provides supplementary
downstream results and evaluation settings, including
the use of multiple data seeds or repeated evaluation runs
where applicable.
These details support replication of our experimental procedures and results.

\bibliography{iclr2027_conference}
\bibliographystyle{iclr2027_conference}

\clearpage
\appendix

\section{Experimental Setup and Reproducibility}
\label{app:setup}
\subsection{Details about LongPPL and ShortPPL}
\label{app:ppl_metric}
This section supplements the metric definitions in the main text and records statistics of the test set used for all PPL evaluations. LongPPL and ShortPPL use the same GovReport documents~\citep{huang2021efficient}, reference model,
and full-context forward pass; they differ only in the token positions over
which losses are aggregated.

\paragraph{Token selection.}
We precompute token annotations with Llama-3.1-8B and its tokenizer and keep them fixed for every evaluated model. As in the main text, let
\[
\Delta_t =
\ell_{\mathrm{ref}}^{\mathrm{chunk}}(x_t)
-
\ell_{\mathrm{ref}}^{\mathrm{full}}(x_t),
\]

where the local context contains 4,096 tokens. LongPPL selects positions that are easy for the reference model under full context but become substantially harder without distant context, whereas ShortPPL selects a complementary set of positions that are also easy under full context but have lower loss with only local context:
\[
\begin{aligned}
\mathcal K_{\mathrm{long}}
&=
\left\{t:
\ell_{\mathrm{ref}}^{\mathrm{full}}(x_t)<2,\;
\Delta_t>2
\right\},\\
\mathcal K_{\mathrm{short}}
&=
\left\{t:
\ell_{\mathrm{ref}}^{\mathrm{full}}(x_t)<2,\;
\Delta_t<0
\right\}.
\end{aligned}
\]

The two sets are therefore disjoint. At evaluation time, each document is decoded to text, retokenized with the evaluated model's tokenizer, and the fixed character intervals are mapped to that model's token positions. This permits the same text spans to be used when the evaluated tokenizer differs from the reference tokenizer.

\paragraph{Test-set construction.}
The test set contains 10,000 GovReport documents. Candidate documents contain 32,768--39,999 tokens under the Llama-3.1 tokenizer; each document is then retokenized with the evaluated model's tokenizer and truncated to at most 32,768 tokens. Because the tokenizers have different granularities, a small fraction of the retokenized sequences are shorter than this cap. In the final evaluation inputs, 9,434 sequences contain exactly 32,768 tokens; across all 10,000 sequences, the mean and median lengths are 32,751 and 32,768 tokens, respectively.

\begin{table*}[htbp]
    \centering
    \caption{Statistics of the GovReport PPL test set. Document length is measured with the Llama-3.1 tokenizer; the remaining rows use the tokenizer shared by our evaluated models after truncation at 32,768 tokens.}
    \label{tab:ppl_test_statistics}
    \small
    \setlength{\tabcolsep}{5pt}
    \begin{tabular}{lrrrrr}
        \toprule
        Quantity & Samples & Total tokens & Mean & Median & Min--max \\
        \midrule
        Source document length & 10,000 & 362,174,270 & 36,217 & 36,121 & 32,768--39,999 \\
        Evaluated sequence length & 10,000 & 327,512,254 & 32,751 & 32,768 & 24,709--32,768 \\
        Long key tokens & 10,000 & 4,662,555 & 466.3 & 330 & 6--7,127 \\
        Short key tokens & 10,000 & 82,542,674 & 8,254.3 & 8,137 & 1,978--18,737 \\
        \bottomrule
    \end{tabular}
\end{table*}

As shown in Table~\ref{tab:ppl_test_statistics}, Long key tokens account for 1.42\% of all evaluated tokens, compared with 25.20\% for Short key tokens.
Because the number of selected tokens varies substantially across documents, we compute both metrics by averaging negative log-likelihoods over all selected tokens in the test set and then exponentiating, rather than weighting documents equally.

\subsection{Alternative Scaling-Law Formulations}
\label{app:formula}
We compare Eq.~\ref{eq:lccp_scaling_law} with alternative offset constraints, data-aggregation assumptions, and interaction terms on both the \textsc{Linear-Decay} and \textsc{Constant-Last4} data.

Within each learning-rate setting, every log-space variant uses the same data, held-out-scale splits, sample weighting, exponent regularization, and multi-start optimization as the corresponding main result: S4 is predicted from S1--S3, while S5 is predicted from S1--S4.

For scaling-law evaluation, let $y_i=\log(\mathrm{LongPPL}_i)$ and $\hat y_i$ denote its prediction. We report $R^2=1-\sum_i(\hat y_i-y_i)^2/\sum_i(y_i-\bar y)^2$ and $\mathrm{MAPE}=\frac{100}{n}\sum_i|\hat y_i-y_i|/|y_i|$, with MAPE expressed as a percentage.
For numerical conditioning, define $\widetilde N=N/10^8$, $\widetilde D_s=D_s/10^9$, and $\widetilde D_l=D_l/10^9$.
These normalizations are absorbed into the fitted amplitudes and do not change the function class of any formula.
For compactness, Tables~\ref{tab:formula_ablation} and \ref{tab:formula_ablation_constant} report only the held-out scale-extrapolation metrics.
As a diagnostic, across both training splits and all formulas, the unweighted training-set $R^2$ ranges from $0.9633$ to $0.9943$ on the \textsc{Linear-Decay} and from $0.9740$ to $0.9954$ on the \textsc{Constant-Last4} data.
The lower training-fit values arise mainly from the combined-budget and effective-data variants.
These results show that training fit alone does not adequately distinguish the predictive performance of different formulations. We therefore emphasize held-out scale extrapolation when evaluating the proposed law and its alternatives.
\begin{table*}[t]
    \centering
    \caption{Scale-extrapolation results for formula ablations on the \textsc{Linear-Decay} data. Both $R^2$ and MAPE are computed in $\log(\mathrm{LongPPL})$ space. S5 contains five diagonal and four off-diagonal budget configurations. For the three powered-pair variants, $k$ is allowed to vary over $0<k\leq2$.}
    \label{tab:formula_ablation}
    \small
    \setlength{\tabcolsep}{3pt}
    \renewcommand{\arraystretch}{1.4}
    \resizebox{\textwidth}{!}{%
    \begin{tabular}{llcccc}
        \toprule
        Variant & Full formula & \multicolumn{2}{c}{S4} & \multicolumn{2}{c}{S5} \\
        \cmidrule(lr){3-4} \cmidrule(lr){5-6}
        & & $R^2$ & MAPE & $R^2$ & MAPE \\
        \midrule
        No offset & $\log(\mathrm{LongPPL})=\frac{A}{N^\alpha}+\frac{B_s}{D_s^{\beta_s}}+\frac{B_l}{D_l^{\beta_l}}$ & 0.7995 & 3.62\% & -0.1031 & 9.05\% \\
        Nonnegative offset & $\log(\mathrm{LongPPL})=\frac{A}{N^\alpha}+\frac{B_s}{D_s^{\beta_s}}+\frac{B_l}{D_l^{\beta_l}}+C,\ C\geq0$ & 0.7995 & 3.62\% & -0.1031 & 9.05\% \\
        Free offset (ours) & $\log(\mathrm{LongPPL})=\frac{A}{N^\alpha}+\frac{B_s}{D_s^{\beta_s}}+\frac{B_l}{D_l^{\beta_l}}+C$ & \textbf{0.9663} & \textbf{1.43\%} & \textbf{0.9516} & \underline{1.71\%} \\
        \midrule
        Combined budget & $\log(\mathrm{LongPPL})=\frac{A}{N^\alpha}+\frac{B}{(D_s+D_l)^\beta}+C$ & 0.8935 & 2.53\% & 0.8494 & 2.40\% \\
        Effective data & $\log(\mathrm{LongPPL})=\frac{A}{N^\alpha}+\frac{B}{(D_l+KD_s)^\beta}+C$ & \underline{0.9079} & \underline{2.14\%} & 0.8280 & 2.94\% \\
        Scale-dependent effective data & $\log(\mathrm{LongPPL})=\frac{A}{N^\alpha}+\frac{B}{(D_l+KN^\nu D_s)^\beta}+C$ & 0.0980 & 7.68\% & 0.8346 & 3.19\% \\
        \midrule
        Powered pair ($N$--$D_s$) & $\log(\mathrm{LongPPL})=\left(\frac{A}{N^\alpha}+\frac{B_s}{D_s^{\beta_s}}\right)^k+\frac{B_l}{D_l^{\beta_l}}+C,\ 0<k\leq2$ & 0.5311 & 5.44\% & 0.7990 & 2.82\% \\
        Powered pair ($N$--$D_l$) & $\log(\mathrm{LongPPL})=\left(\frac{A}{N^\alpha}+\frac{B_l}{D_l^{\beta_l}}\right)^k+\frac{B_s}{D_s^{\beta_s}}+C,\ 0<k\leq2$ & 0.4370 & 6.37\% & 0.4530 & 5.22\% \\
        Powered pair ($D_s$--$D_l$) & $\log(\mathrm{LongPPL})=\frac{A}{N^\alpha}+\left(\frac{B_s}{D_s^{\beta_s}}+\frac{B_l}{D_l^{\beta_l}}\right)^k+C,\ 0<k\leq2$ & 0.7984 & 3.58\% & \underline{0.9488} & \textbf{1.57\%} \\
        CPT interaction & $\log(\mathrm{LongPPL})=\frac{A}{N^\alpha}+\frac{B_s}{D_s^{\beta_s}}+\frac{B_l}{D_l^{\beta_l}N^\eta}+C$ & 0.7557 & 4.14\% & 0.7163 & 3.50\% \\
        \bottomrule
    \end{tabular}%
    }
\end{table*}

\begin{table*}[t]
    \centering
    \caption{The same formula ablations on the \textsc{Constant-Last4} data, with both $R^2$ and MAPE computed in $\log(\mathrm{LongPPL})$ space. S4 contains a $5\times5$ grid, whereas S5 contains 24 points that hold $N$ and $D_s$ fixed and vary only $D_l$. For the three powered-pair variants, $k$ is allowed to vary over $0<k\leq2$.}
    \label{tab:formula_ablation_constant}
    \small
    \setlength{\tabcolsep}{3pt}
    \renewcommand{\arraystretch}{1.4}
    \resizebox{\textwidth}{!}{%
    \begin{tabular}{llcccc}
        \toprule
        Variant & Full formula & \multicolumn{2}{c}{S4} & \multicolumn{2}{c}{S5} \\
        \cmidrule(lr){3-4} \cmidrule(lr){5-6}
        & & $R^2$ & MAPE & $R^2$ & MAPE \\
        \midrule
        No offset & $\log(\mathrm{LongPPL})=\frac{A}{N^\alpha}+\frac{B_s}{D_s^{\beta_s}}+\frac{B_l}{D_l^{\beta_l}}$ & 0.9354 & 1.89\% & -4.0917 & 11.01\% \\
        Nonnegative offset & $\log(\mathrm{LongPPL})=\frac{A}{N^\alpha}+\frac{B_s}{D_s^{\beta_s}}+\frac{B_l}{D_l^{\beta_l}}+C,\ C\geq0$ & 0.9354 & 1.89\% & -4.0917 & 11.01\% \\
        Free offset (ours) & $\log(\mathrm{LongPPL})=\frac{A}{N^\alpha}+\frac{B_s}{D_s^{\beta_s}}+\frac{B_l}{D_l^{\beta_l}}+C$ & \textbf{0.9747} & \textbf{1.08\%} & \textbf{0.9718} & \textbf{0.66\%} \\
        \midrule
        Combined budget & $\log(\mathrm{LongPPL})=\frac{A}{N^\alpha}+\frac{B}{(D_s+D_l)^\beta}+C$ & 0.8901 & 1.95\% & 0.1388 & 4.02\% \\
        Effective data & $\log(\mathrm{LongPPL})=\frac{A}{N^\alpha}+\frac{B}{(D_l+KD_s)^\beta}+C$ & 0.9155 & \underline{1.46\%} & \underline{0.7943} & \underline{1.49\%} \\
        Scale-dependent effective data & $\log(\mathrm{LongPPL})=\frac{A}{N^\alpha}+\frac{B}{(D_l+KN^\nu D_s)^\beta}+C$ & 0.4727 & 5.45\% & -0.7878 & 6.20\% \\
        \midrule
        Powered pair ($N$--$D_s$) & $\log(\mathrm{LongPPL})=\left(\frac{A}{N^\alpha}+\frac{B_s}{D_s^{\beta_s}}\right)^k+\frac{B_l}{D_l^{\beta_l}}+C,\ 0<k\leq2$ & 0.8190 & 3.20\% & -0.3086 & 5.49\% \\
        Powered pair ($N$--$D_l$) & $\log(\mathrm{LongPPL})=\left(\frac{A}{N^\alpha}+\frac{B_l}{D_l^{\beta_l}}\right)^k+\frac{B_s}{D_s^{\beta_s}}+C,\ 0<k\leq2$ & 0.7268 & 4.19\% & -2.3259 & 8.86\% \\
        Powered pair ($D_s$--$D_l$) & $\log(\mathrm{LongPPL})=\frac{A}{N^\alpha}+\left(\frac{B_s}{D_s^{\beta_s}}+\frac{B_l}{D_l^{\beta_l}}\right)^k+C,\ 0<k\leq2$ & 0.9454 & 1.61\% & 0.0805 & 4.58\% \\
        CPT interaction & $\log(\mathrm{LongPPL})=\frac{A}{N^\alpha}+\frac{B_s}{D_s^{\beta_s}}+\frac{B_l}{D_l^{\beta_l}N^\eta}+C$ & \underline{0.9537} & 1.59\% & -0.2800 & 5.34\% \\
        \bottomrule
    \end{tabular}%
    }
\end{table*}

\paragraph{Offset constraints.}
A zero-offset additive law, as used in some prior scaling analyses~\citep{qiao2026rethinking}, has a plausible asymptotic interpretation: loss may approach zero as model and data scales grow, especially for the chosen predictable key tokens in LongPPL.
It is nevertheless substantially less accurate over the finite experimental range under both learning-rate settings.
For both settings, constraining $C\geq0$ gives the same result as the zero-offset law because the optimizer drives $C$ to numerical zero.
The fitted values are $1.6\times10^{-48}$ and $4.6\times10^{-37}$ for the \textsc{Linear-Decay} S1--S3 and S1--S4 fits, respectively, and $2.5\times10^{-49}$ and
$2.0\times10^{-36}$ for their \textsc{Constant-Last4} counterparts.
Allowing $C$ to be free instead yields $C=-0.60$ and $C=-0.82$ for the \textsc{Linear-Decay} S1--S3 and S1--S4 fits, respectively, and $C=-0.71$ and $C=-0.61$ for the corresponding \textsc{Constant-Last4} fits.

\begin{figure*}[t]
\centering
\includegraphics[width=\textwidth]{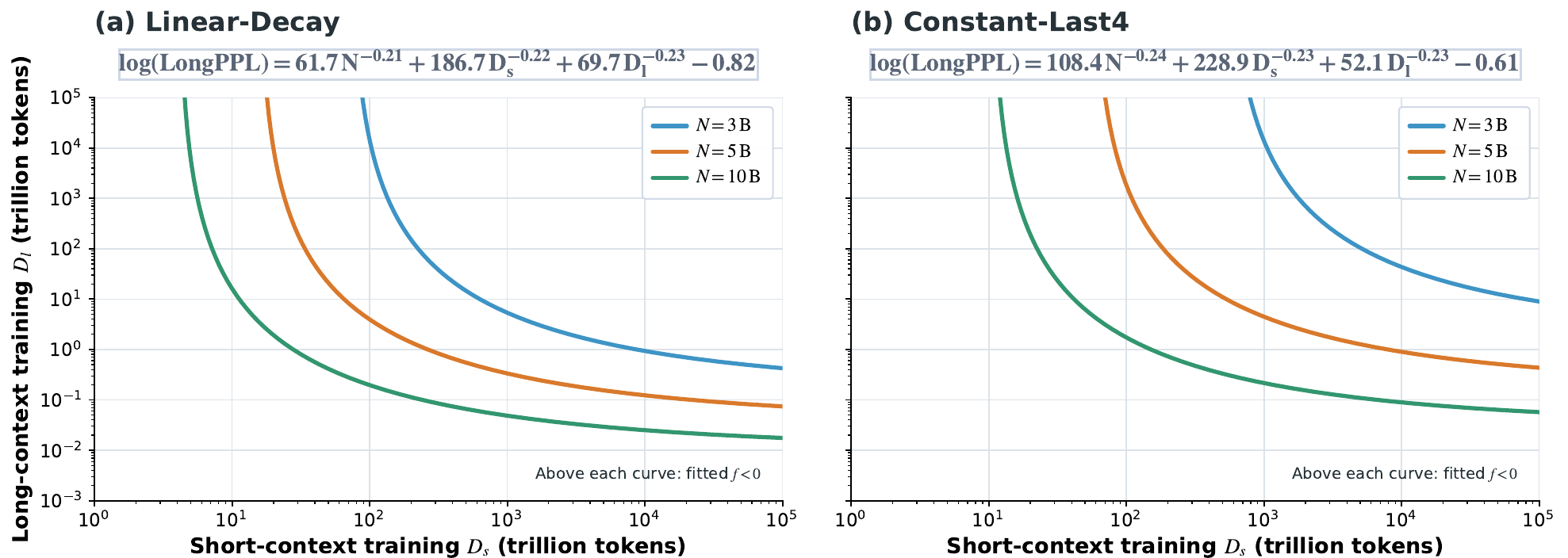}
\caption{
Zero contours of the free-offset additive scaling law fitted on S1--S4. The left and right panels show the \textsc{Linear-Decay} and \textsc{Constant-Last4} fits, respectively. Each curve contains the $(D_s,D_l)$ combinations for which the fitted
$\log(\mathrm{LongPPL})$ equals zero at the indicated non-embedding parameter count $N$.}
\label{fig:zero_contours}
\end{figure*}

Importantly, a negative offset alone does not imply negative predictions within the relevant scaling regime, since the model- and data-dependent terms remain positive. 
To stress-test this behavior, Figure~\ref{fig:zero_contours} plots the zero contours of the S1--S4 fits at several extrapolated model sizes.
Finite zero contours emerge only at larger model scales and extreme data budgets. For example, at $N=10$B, setting $D_s=D_l$ gives a zero crossing at approximately $11.0$T tokens per stage for \textsc{Linear-Decay} and $29.5$T tokens per stage for \textsc{Constant-Last4}. These budgets are orders of magnitude beyond the S1--S4 fitting range, whose maxima are approximately $0.11$T short-context tokens and $0.012$T long-context tokens. Predictions below zero imply $\mathrm{LongPPL}<1$, indicating that the fitted law has been extrapolated beyond its valid range. The negative offset should therefore be interpreted as a finite-range correction rather than an asymptotic loss floor.

\paragraph{Combining the two training stages.}
We test whether short-context pre-training and LCCP data can be collapsed into a single data variable using three formulations. The combined-budget baseline, $D_s+D_l$, assumes that the two types of tokens are interchangeable. The effective-data formulation, $D_l+KD_s$, learns a fixed conversion factor between them, while the scale-dependent formulation, $D_l+K\widetilde N^\nu D_s$, allows this conversion factor to vary with model scale.

All three formulations are less accurate than our additive law, which models $D_s$ and $D_l$ separately. On the linear-decay data, our law achieves $R^2_{\log}=0.9663$ on S4 and $0.9516$ on S5, compared with $0.8935$ and $0.8494$ for the combined-budget baseline. The effective-data formulation improves some results but remains inferior overall, while the scale-dependent formulation is unstable across extrapolation settings. The same conclusion holds for the constant-learning-rate data, where our law achieves $R^2_{\log}=0.9747$ on S4 and $0.9718$ on S5 and outperforms all three alternatives.

\paragraph{Interaction terms.}
We test two types of interaction terms to examine whether the additive law overlooks coupling among the scaling variables.
First, motivated by a previously proposed coupled form~\citep{videau2026skalingchinchillasexponentsmeet}, we replace each pairwise sum of the terms associated with $N$, $D_s$, and $D_l$ by $(x_i+x_j)^k$ in turn, while leaving the third term additive. The three resulting formulas are given explicitly in Tables~\ref{tab:formula_ablation} and \ref{tab:formula_ablation_constant}.
We constrain the coupling exponent to $0<k\le2$. Across both datasets, both held-out scales, and all three pairings, the fitted value reaches the upper bound to numerical precision. The tested constrained fits provide no consistent extrapolation advantage over the additive law.

Second, following the motivation of prior work on continued pretraining~\citep{zheng2024breakinglanguagebarrierscrosslingual}, we consider the interaction
$C+A/N^\alpha+B_s/D_s^{\beta_s}+B_l/(D_l^{\beta_l}N^\eta)$. This formulation allows model size to modulate the contribution of LCCP data to long-context learning.
This interaction is less accurate than the free-offset additive law in all four extrapolation tests. On the linear-decay data, it achieves $R^2_{\log}=0.7557$ on S4 and $0.7163$ on S5. On the constant-learning-rate data, it remains relatively accurate on S4, with $R^2_{\log}=0.9537$, but deteriorates to $R^2_{\log}=-0.2800$ on S5.

Overall, the tested interaction parameterizations provide no consistent extrapolation benefit. This does not rule out interactions among the scaling variables, but indicates that these particular forms do not outperform the simpler additive law over the measured range.


\subsection{Training Details}
\label{app:training}
This section records the complete training procedure used for the S1--S5 scaling experiments and for the 128K extension and supervised fine-tuning experiments in Sections~\ref{sec:128k} and~\ref{sec:sft}. We report realized training token counts, rather than only their nominal multiples of $N$, since training budgets must be rounded to optimizer steps.
Below, $N$ denotes the non-embedding parameter count in Table~\ref{tab:model_configs}, and GBS/MBS denote the global and per-device micro batch sizes in sequences.

\paragraph{Common settings.}
All stages use the Muon optimizer, bfloat16 training, weight decay $0.1$, and gradient clipping at $1.0$. Attention and hidden-state dropout are both zero.
The 4K, 32K, and 128K stages use RoPE base frequencies of $10{,}000$, $500{,}000$, and $5{,}000{,}000$, respectively, and the 128K stage uses full activation recomputation.
All runs were performed on two nodes of eight NVIDIA H800 GPUs each (16 GPUs in total).

\paragraph{4K short-context pre-training.}
For every model scale, we train five WSD runs with realized budgets close to $90N$, $180N$, $270N$, $360N$, and $450N$. Each run consists of a linear warmup, a constant-learning-rate stable phase, and a linear decay phase; 
the warmup-plus-stable portion and the decay portion contain tokens in an $8{:}1$ ratio, and $D_s$ counts both. The decay phase reduces the learning rate linearly from $\eta_{\mathrm{stable}}$ to $\eta_{\mathrm{end}}=\eta_{\mathrm{stable}}/10$;
this terminal value is also the starting learning rate of the subsequent 32K stage. Warmup lasts 1,000 optimizer steps at every scale except S1, where we use 500: at S1 a 1,000-step warmup would consume $0.52$B tokens, or $18.5\%$ of the
smallest ($90N$) budget, whereas 500 steps gives a warmup fraction of $9.3\%$, comparable to the $9.6\%$ at S2.  Table~\ref{tab:short_training_details} lists the realized budgets.

\begin{table*}[t]
    \centering
    \caption{Short-context pre-training configurations. All runs use a 4,096-token
    sequence length and RoPE base $10{,}000$. Token counts are in billions and
    correspond, from left to right, to the nominal $90N$, $180N$, $270N$, $360N$,
    and $450N$ budgets.}
    \label{tab:short_training_details}
    \scriptsize
    \setlength{\tabcolsep}{3.2pt}
    \renewcommand{\arraystretch}{1.2}
    \resizebox{\textwidth}{!}{%
    \begin{tabular}{lrrrrrrrrrrr}
        \toprule
        \multirow{2}{*}{Scale} & \multirow{2}{*}{$N$ (M)} &
        \multirow{2}{*}{$\eta_{\mathrm{stable}}$} &
        \multirow{2}{*}{$\eta_{\mathrm{end}}$} &
        \multirow{2}{*}{GBS} & \multirow{2}{*}{MBS} &
        \multirow{2}{*}{Warmup} & \multicolumn{5}{c}{Realized $D_s$ (B)} \\
        \cmidrule(lr){8-12}
        & & & & & & & $90N$ & $180N$ & $270N$ & $360N$ & $450N$ \\
        \midrule
        S1 & 31.470  & $9.7656\!\times\!10^{-4}$ & $9.766\!\times\!10^{-5}$ & 128 & 8 & 500   & 2.83 & 5.66 & 8.49 & 11.32 & 14.16 \\
        S2 & 60.845  & $9.5419\!\times\!10^{-4}$ & $9.542\!\times\!10^{-5}$ & 128 & 8 & 1,000 & 5.48 & 10.96 & 16.43 & 21.91 & 27.39 \\
        S3 & 104.427 & $9.7656\!\times\!10^{-4}$ & $9.766\!\times\!10^{-5}$ & 256 & 8 & 1,000 & 9.40 & 18.80 & 28.20 & 37.59 & 46.99 \\
        S4 & 245.417 & $9.7656\!\times\!10^{-4}$ & $9.766\!\times\!10^{-5}$ & 256 & 4 & 1,000 & 22.08 & 44.17 & 66.25 & 88.33 & 110.42 \\
        S5 & 821.147 & $9.7656\!\times\!10^{-4}$ & $9.766\!\times\!10^{-5}$ & 512 & 4 & 1,000 & 73.99 & 147.98 & 221.96 & 295.95 & 369.94 \\
        \bottomrule
    \end{tabular}%
    }
\end{table*}

\paragraph{32K long-context continual pre-training.}
\label{app:32k_lccp}
When increasing the sequence length from 4K to 32K, we reduce the global batch size by a factor of eight so that the number of tokens per optimizer step is unchanged. Every 32K run uses MBS 1 and a RoPE base $500{,}000$. The \textsc{Linear-Decay} runs have no warmup, begin at the terminal learning rate of 4K pre-training, and linearly decay to one tenth of that rate. For S1--S4,
we train the full $5\times5$ grid of short- and long-context budgets; for S5,
we train the nine diagonal and off-diagonal configurations described in Section~\ref{sec:scaling_law}.

The \textsc{Constant-Raw} trajectories use the same batch sizes, data mixture, and starting learning rates as the \textsc{Linear-Decay} runs, but keep the learning rate constant. For S1, S2, S4, and S5, the trajectories are initialized from the same short-context checkpoints as their corresponding \textsc{Linear-Decay} runs. For S3, we retrained the short-context models because the original checkpoints were no longer available; the replacement runs have slightly different realized short-context training budgets, as reported in Table~\ref{tab:lccp_constant_training_details}. 
S1--S4 are continued through approximately $50N$ of long-context data. The S5 trajectory fixes the nominal short-context budget at $D_s=450N$ and continues LCCP through $D_l=120N$. We evaluate this trajectory every $5N$, starting at $D_l=5N$, with adjacent checkpoints for merging spaced by $1N$. The realized budgets are summarized in Table~\ref{tab:lccp_constant_training_details}. All scaling-law fits use the realized model sizes and token counts of the corresponding runs.

\begin{table*}[t]
    \centering
    \caption{\textsc{Linear-Decay} 32K LCCP configurations. All runs use a 32,768-token sequence length, RoPE base $500{,}000$, and no warmup. Token counts are in billions and correspond, from left to right, to the nominal $10N$, $20N$, $30N$, $40N$, and $50N$ budgets.}
    \label{tab:lccp_training_details}
    \scriptsize
    \setlength{\tabcolsep}{3.2pt}
    \renewcommand{\arraystretch}{1.2}
    \resizebox{\textwidth}{!}{%
    \begin{tabular}{lrrrrrrrrrrr}
        \toprule
        \multirow{2}{*}{Scale} & \multirow{2}{*}{$N$ (M)} &
        \multirow{2}{*}{$\eta_{\mathrm{start}}$} &
        \multirow{2}{*}{$\eta_{\mathrm{end}}$} &
        \multirow{2}{*}{GBS} & \multirow{2}{*}{MBS} &
        \multirow{2}{*}{Warmup} & \multicolumn{5}{c}{Realized $D_l$ (B)} \\
        \cmidrule(lr){8-12}
        & & & & & & & $10N$ & $20N$ & $30N$ & $40N$ & $50N$ \\
        \midrule
        S1 & 31.470  & $9.766\!\times\!10^{-5}$ & $9.766\!\times\!10^{-6}$ & 16 & 1 & 0 & 0.31 & 0.63 & 0.94 & 1.26 & 1.57 \\
        S2 & 60.845  & $9.542\!\times\!10^{-5}$ & $9.542\!\times\!10^{-6}$ & 16 & 1 & 0 & 0.61 & 1.22 & 1.83 & 2.43 & 3.04 \\
        S3 & 104.427 & $9.766\!\times\!10^{-5}$ & $9.766\!\times\!10^{-6}$ & 32 & 1 & 0 & 1.04 & 2.09 & 3.13 & 4.18 & 5.22 \\
        S4 & 245.417 & $9.766\!\times\!10^{-5}$ & $9.766\!\times\!10^{-6}$ & 32 & 1 & 0 & 2.45 & 4.91 & 7.36 & 9.81 & 12.27 \\
        S5 & 821.147 & $9.766\!\times\!10^{-5}$ & $9.766\!\times\!10^{-6}$ & 64 & 1 & 0 & 8.22 & 16.44 & 24.66 & 32.88 & 41.10 \\
        \bottomrule
    \end{tabular}%
    }
\end{table*}

\begin{table*}[t]
    \centering
    \caption{\textsc{Constant-Raw}/\textsc{Constant-Last4} 32K LCCP
    configurations. Each row reports the realized short- and long-context token
    counts. All runs use a 32,768-token sequence
    length, RoPE base $500{,}000$, and no warmup. The identical start and end
    learning rates indicate a constant schedule. Token counts are in billions;
    $\backslash$ denotes an unavailable configuration.}
    \label{tab:lccp_constant_training_details}
    \scriptsize
    \setlength{\tabcolsep}{2.2pt}
    \renewcommand{\arraystretch}{1.2}
    \resizebox{\textwidth}{!}{%
    \begin{tabular}{lrrrrrrrrrrrrr}
        \toprule
        \multirow{2}{*}{Scale} & \multirow{2}{*}{$N$ (M)} &
        \multirow{2}{*}{Realized $D_s$ (B)} &
        \multirow{2}{*}{$\eta_{\mathrm{start}}$} &
        \multirow{2}{*}{$\eta_{\mathrm{end}}$} &
        \multirow{2}{*}{GBS} & \multirow{2}{*}{MBS} &
        \multirow{2}{*}{Warmup} & \multicolumn{6}{c}{Realized $D_l$ (B)} \\
        \cmidrule(lr){9-14}
        & & & & & & & & $10N$ & $20N$ & $30N$ & $40N$ & $50N$ & $120N$ \\
        \midrule
        \multirow{5}{*}{S1} & \multirow{5}{*}{31.470} & 2.83 & \multirow{5}{*}{$9.766\!\times\!10^{-5}$} & \multirow{5}{*}{$9.766\!\times\!10^{-5}$} & \multirow{5}{*}{16} & \multirow{5}{*}{1} & \multirow{5}{*}{0} & 0.31 & 0.63 & 0.94 & 1.26 & 1.57 & $\backslash$ \\
        & & 5.66 & & & & & & 0.31 & 0.63 & 0.94 & 1.26 & 1.57 & $\backslash$ \\
        & & 8.49 & & & & & & 0.31 & 0.63 & 0.94 & 1.26 & 1.57 & $\backslash$ \\
        & & 11.32 & & & & & & 0.31 & 0.63 & 0.94 & 1.26 & 1.57 & $\backslash$ \\
        & & 14.16 & & & & & & 0.31 & 0.63 & 0.94 & 1.26 & 1.57 & $\backslash$ \\
        \midrule
        \multirow{5}{*}{S2} & \multirow{5}{*}{60.845} & 5.48 & \multirow{5}{*}{$9.542\!\times\!10^{-5}$} & \multirow{5}{*}{$9.542\!\times\!10^{-5}$} & \multirow{5}{*}{16} & \multirow{5}{*}{1} & \multirow{5}{*}{0} & 0.60 & 1.21 & 1.82 & 2.43 & 3.04 & $\backslash$ \\
        & & 10.96 & & & & & & 0.60 & 1.21 & 1.82 & 2.42 & 3.04 & $\backslash$ \\
        & & 16.43 & & & & & & 0.59 & 1.20 & 1.81 & 2.42 & 3.04 & $\backslash$ \\
        & & 21.91 & & & & & & 0.59 & 1.20 & 1.81 & 2.41 & 3.04 & $\backslash$ \\
        & & 27.39 & & & & & & 0.58 & 1.19 & 1.80 & 2.41 & 3.04 & $\backslash$ \\
        \midrule
        \multirow{5}{*}{S3} & \multirow{5}{*}{104.427} & 9.89 & \multirow{5}{*}{$9.766\!\times\!10^{-5}$} & \multirow{5}{*}{$9.766\!\times\!10^{-5}$} & \multirow{5}{*}{32} & \multirow{5}{*}{1} & \multirow{5}{*}{0} & 1.12 & 2.22 & 3.32 & 4.42 & 5.49 & $\backslash$ \\
        & & 19.78 & & & & & & 1.14 & 2.24 & 3.34 & 4.44 & 5.49 & $\backslash$ \\
        & & 29.67 & & & & & & 1.16 & 2.26 & 3.36 & 4.46 & 5.49 & $\backslash$ \\
        & & 39.56 & & & & & & 1.18 & 2.28 & 3.38 & 4.48 & 5.49 & $\backslash$ \\
        & & 49.45 & & & & & & 1.20 & 2.30 & 3.40 & 4.50 & 5.49 & $\backslash$ \\
        \midrule
        \multirow{5}{*}{S4} & \multirow{5}{*}{245.417} & 22.08 & \multirow{5}{*}{$9.766\!\times\!10^{-5}$} & \multirow{5}{*}{$9.766\!\times\!10^{-5}$} & \multirow{5}{*}{32} & \multirow{5}{*}{1} & \multirow{5}{*}{0} & 2.45 & 4.91 & 7.36 & 9.81 & 12.27 & $\backslash$ \\
        & & 44.17 & & & & & & 2.45 & 4.91 & 7.36 & 9.81 & 12.27 & $\backslash$ \\
        & & 66.25 & & & & & & 2.45 & 4.91 & 7.36 & 9.81 & 12.27 & $\backslash$ \\
        & & 88.33 & & & & & & 2.45 & 4.91 & 7.36 & 9.81 & 12.27 & $\backslash$ \\
        & & 110.42 & & & & & & 2.45 & 4.91 & 7.36 & 9.81 & 12.27 & $\backslash$ \\
        \midrule
        S5 & 821.147 & 369.94 & $9.766\!\times\!10^{-5}$ & $9.766\!\times\!10^{-5}$ & 64 & 1 & 0 & 8.22 & 16.44 & 24.66 & 32.88 & 41.10 & 98.65 \\
        \bottomrule
    \end{tabular}%
    }
\end{table*}

\paragraph{128K context extension and supervised fine-tuning.}
For the experiment in Section~\ref{sec:128k}, we apply the same 128K extension recipe to the selected S5 \textsc{Constant-Raw} and \textsc{Constant-Last4} checkpoints. We increase the RoPE base to $5{,}000{,}000$ and use a data mixture with a larger fraction of sequences longer than 64K. For long-context SFT in Section~\ref{sec:sft}, every selected checkpoint undergoes the same 2.10B-token training stage on a mixture of general instruction-following examples and high-quality long-context question-answering data. Because agentic SFT requires a larger capability shift from the preceding training stages, we double the training budget to 4.20B tokens while retaining the same linear-decay learning-rate schedule. All three stages use span-based attention to isolate packed documents. Table~\ref{tab:post_lccp_training_details} provides the remaining settings.
\begin{table}[t]
    \centering
    \caption{Training configurations applied after 32K LCCP. The token counts
    are the realized values obtained from the listed sequence length, GBS, and
    number of optimizer steps.}
    \label{tab:post_lccp_training_details}
    \scriptsize
    \setlength{\tabcolsep}{3pt}
    \resizebox{\linewidth}{!}{%
    \begin{tabular}{lrrrrlll}
        \toprule
        Stage & Seq. len. & Steps & GBS/MBS & Tokens (B) & LR schedule & RoPE base \\
        \midrule
        128K extension & 131,072 & 1000 & 16/1 & 2.10 & linear: $9.766\!\times\!10^{-5}\!\rightarrow\!9.766\!\times\!10^{-7}$ & $5{,}000{,}000$ \\
        Long-context SFT & 32,768 & 2000 & 32/1 & 2.10 & linear: $9.766\!\times\!10^{-5}\!\rightarrow\!9.766\!\times\!10^{-7}$ & $500{,}000$ \\
        Agentic SFT & 32,768 & 4000 & 32/1 & 4.20 & linear: $9.766\!\times\!10^{-5}\!\rightarrow\!9.766\!\times\!10^{-7}$ & $500{,}000$ \\
        \bottomrule
    \end{tabular}%
    }
\end{table}

\subsection{Training Data}
\label{app:training-data}

\paragraph{LCCP data.}
The LCCP corpus combines publicly available and internally curated data, primarily covering Chinese and English. Public sources include filtered web text, such as FineWeb- and HPLT-derived collections~\citep{penedo2024fineweb, wang2025ultra, de2024new}; books and long-form documents; academic papers and PDF text; legal, financial, and patent documents; source code and technical documentation, such as data derived from RefineCode and Stack v2~\citep{huang2025opencoder,lozhkov2024starcoder}; and publicly released synthetic and instruction data from the Nemotron family~\citep{bercovich2025llama, NemotronPostTrainingDatasetV1}. The same 32K mixture is used across all model scales and training-budget configurations.

\paragraph{Length-aware LCCP mixtures.}
Documents are grouped by their tokenized source-document lengths before sequence construction. Table~\ref{tab:lccp_length_mixture} reports the configured sampling mass of each length bucket. These values describe the source-document lengths, rather than the lengths of the final training sequences, which are fixed at 32,768 and 131,072 tokens for the 32K and 128K
stages, respectively.

In the 32K mixture, documents shorter than 16K tokens account for approximately 52.57\% of the sampling mass, while documents shorter than 32K tokens account for approximately 65.11\%. Documents of at least 32K tokens account for approximately 34.89\%. For the 128K extension, the same broad source pool is retained while more sampling mass is assigned to longer documents. 
The share of documents shorter than 16K tokens decreases to approximately 40.60\%, and the share of documents shorter than 32K tokens decreases to 50.00\%. Accordingly, the share of documents of at least 32K tokens increases to 50.00\%. More specifically, the share of documents in the 32K--128K range increases from 19.80\% to 28.38\%, while the share of
documents of at least 128K tokens increases from 15.09\% to 21.62\%.

\begin{table*}[t]
\centering
\caption{Aggregate LCCP sampling weights by tokenized source-document length. Values denote percentages of the configured sampling mass.}
\label{tab:lccp_length_mixture}
\scriptsize
\setlength{\tabcolsep}{5pt}
\renewcommand{\arraystretch}{1.15}
\resizebox{0.6\textwidth}{!}{%
\begin{tabular}{lrr}
\toprule
Source-document length & 32K LCCP & 128K extension \\
\midrule
$[0,4\mathrm{K})$               & 27.63\% & 21.90\% \\
$[4\mathrm{K},8\mathrm{K})$     & 14.39\% & 10.79\% \\
$[8\mathrm{K},16\mathrm{K})$    & 10.55\% &  7.91\% \\
$[16\mathrm{K},32\mathrm{K})$   & 12.54\% &  9.40\% \\
$[32\mathrm{K},64\mathrm{K})$   & 11.12\% & 15.94\% \\
$[64\mathrm{K},128\mathrm{K})$  &  8.68\% & 12.44\% \\
$[128\mathrm{K},256\mathrm{K})$ &  7.37\% & 10.56\% \\
$[256\mathrm{K},512\mathrm{K})$ &  4.98\% &  7.13\% \\
$[512\mathrm{K},\infty)$        &  2.74\% &  3.93\% \\
\midrule
Total                            & 100.00\% & 100.00\% \\
\bottomrule
\end{tabular}%
}
\end{table*}

\paragraph{LCCP data processing.}
Documents longer than the target sequence length are split into consecutive segments of the target length, starting from the beginning of each document, with a one-token overlap between adjacent segments. 
All resulting segments are retained rather than selecting a random window. Instruction-form examples that exceed the target length are filtered instead of segmented to preserve their supervision structure. The 32K LCCP stage uses standard causal attention without document-boundary masking, whereas the 128K extension uses span-based attention to isolate packed documents.
We perform exact document-level matching between the LCCP training corpus and the GovReport documents used for LongPPL evaluation, and find no exact matches. This check is limited to exact document overlap.

\paragraph{Supervised fine-tuning data.}
The SFT mixture is fixed across all evaluated initializations and training budgets, including checkpoints obtained from different 32K LCCP budgets and the two checkpoint-selection strategies. 
For long-context SFT, we use the non-thinking response-format variant. Table~\ref{tab:sft_task_mixture} reports the aggregate task-family composition. 
The long-context component consists of history-based question answering (11.52\%), multi-hop reasoning (2.24\%), structured retrieval tasks (4.84\%), and additional long-context reasoning data (21.40\%).
For agentic SFT, we use the thinking response-format variant. The mixture consists of 60\% search-agent trajectories, in which the model interleaves reasoning with calls to a search engine, a web-page extractor, and a Python sandbox, and 40\% function and tool calling trajectories, ranging from single-step tool selection to long-horizon multi-step tool chaining. The loss is computed only on assistant turns. 
SFT examples are packed into sequences of 32,768 tokens, with span-based attention preventing attention across packed-example boundaries.

\begin{table*}[t]
\centering
\small
\caption{Task-level composition of the SFT mixture. Percentages denote
configured sampling mass.}
\label{tab:sft_task_mixture}
\setlength{\tabcolsep}{5pt}
\renewcommand{\arraystretch}{1.15}
\begin{tabular}{p{0.5\textwidth}r}
\toprule
Task family & Sampling mass \\
\midrule
General and multilingual instruction following & 10.00\% \\
Long-context question answering and reasoning  & 40.00\% \\
Search-agent trajectories                       & 30.00\% \\
Function and tool calling                       & 20.00\% \\
\midrule
Total                                           & 100.00\% \\
\bottomrule
\end{tabular}
\end{table*}

\subsection{Scaling-Law Fitting Details}
\label{app:fitting}
Eq.~\ref{eq:lccp_scaling_law} contains seven parameters to estimate. Although S1--S4 provide 100 configurations in total, a direct seven-parameter nonlinear fit can be ill-conditioned or nearly non-identifiable: distinct parameter combinations can yield similar predictions over the observed range, making coefficients and exponents sensitive to noise and extrapolation unreliable. We therefore use the experimental-grid structure to stabilize estimation.

At each model scale, the grid varies the short- and long-context
budgets separately. Every fixed $(N,D_s)$ pair has five observations
at distinct $D_l$ values, and every fixed $(N,D_l)$ pair has five
observations at distinct $D_s$ values. With $N$ and $D_s$ fixed, the remaining terms and offset are constant on the $D_l$ slice, so we fit
$y=b_lD_l^{-\beta_l}+c$ to obtain a local estimate of $\beta_l$.
We repeat this over all available slices in the training split and use the median as a training-data-driven reference; the analogous procedure yields the reference for $\beta_s$.
Taking the median limits the effect of a small number of anomalous slices. These slice fits provide initialization and empirical thresholds
for the exponent regularization; they do not replace the final joint fit.

The joint fit uses Huber loss in $\log(\mathrm{LongPPL})$ space,
with equal weight assigned to each training configuration.
We set \texttt{f\_scale=0.05} by pilot fits on the training set
before held-out evaluation.
We apply the one-sided exponent regularizer
\[
\lambda_s\bigl[\max(\beta_s-\bar{\beta}_s,0)\bigr]^2
+
\lambda_l\bigl[\max(\beta_l-\bar{\beta}_l,0)\bigr]^2,
\qquad
\lambda_s=\lambda_l=1.
\]
The thresholds $\bar{\beta}_s$ and $\bar{\beta}_l$ are obtained
from the slice fits for the corresponding training split.
The regularizer penalizes above-threshold exponents without fixing
them or directly constraining the fitted coefficients and offset $C$.

For each held-out-scale experiment, slice-based exponent estimates,
initialization settings, and regularization hyperparameters are
determined exclusively from the training split.
Specifically, the S4 extrapolation uses only S1--S3 for these choices,
whereas the S5 extrapolation uses only S1--S4.
The held-out scale is used only for evaluation.

Additionally, the one-variable experiments in Section~\ref{sec:simple_law} show that an axis exponent can be estimated reliably when the other two variables are fixed and sufficient observations are available along that axis.
For future fits of the same scaling law, the multiple $D_l$ observations on a single \textsc{Constant-Last4} trajectory can first define a reasonable range for $\beta_l$, followed by joint fitting.

Table~\ref{tab:lccp_fitted_parameters} reports the parameters of
$\log(\mathrm{LongPPL})
=AN^{-\alpha}+B_sD_s^{-\beta_s}+B_lD_l^{-\beta_l}+C$,
where $N$ is the non-embedding parameter count and $D_s,D_l$
are token counts, all expressed as absolute counts.
Figures and evaluation metrics use the unrounded parameters.

\begin{table}[t]
\centering
\caption{
Fitted parameters of the joint scaling law. LD denotes \textsc{Linear-Decay}, and CL4 denotes \textsc{Constant-Last4}.
Training scales are shown in parentheses. Fits on S1--S3 predict S4; fits on S1--S4 predict S5. Parameters are rounded to six decimal places.
}
\label{tab:lccp_fitted_parameters}
\small
\setlength{\tabcolsep}{3pt}
\begin{tabular}{lrrrrrrr}
\toprule
Fit
& $A$ & $\alpha$
& $B_s$ & $\beta_s$
& $B_l$ & $\beta_l$
& $C$ \\
\midrule
LD (S1--S3)
& 74.297012 & 0.223528
& 261.977808 & 0.239361
& 71.681123 & 0.235262
& -0.600638 \\
LD (S1--S4)
& 61.738434 & 0.208380
& 186.719700 & 0.220593
& 69.742890 & 0.234577
& -0.817910 \\
CL4 (S1--S3)
& 94.293321 & 0.232590
& 225.443760 & 0.231024
& 50.066563 & 0.223068
& -0.710403 \\
CL4 (S1--S4)
& 108.446353 & 0.244348
& 228.854134 & 0.231342
& 52.141761 & 0.225344
& -0.613778 \\
\bottomrule
\end{tabular}
\end{table}

\section{Model Merging}
\label{app:merging}
We provide implementation details for the checkpoint-merging methods
considered in Section~\ref{sec:merge}.
For the observation-rule comparison in
Figure~\ref{fig:checkpoint_merge}(a), all merging methods use the same
pool of 25 checkpoints from the S4 constant-learning-rate trajectory
with $D_s=450N$ and a target budget of $D_l=50N$, ordered from
the oldest checkpoint $\theta_1$ to the newest checkpoint $\theta_K$.
All methods in this comparison are evaluated on the same first
1,000 evaluation samples.

\paragraph{Agreement with linear decay.}
For Figure~\ref{fig:checkpoint_merge}(b), we measure agreement
using Lin's concordance correlation coefficient (CCC):
\[
\mathrm{CCC}
=
\frac{2s_{xy}}
{s_x^2+s_y^2+(\bar{x}-\bar{y})^2},
\]
where $x_i$ and $y_i$ are the perplexities of the matched
\textsc{Linear-Decay} and constant-learning-rate models,
respectively, across the $n=25$ S4 configurations.
Here, $\bar{x}$ and $\bar{y}$ are sample means;
$s_x^2$, $s_y^2$, and $s_{xy}$ are the variances and covariance
computed with denominator $n$.
CCC is computed separately for ShortPPL and LongPPL,
and for \textsc{Constant-Raw} and \textsc{Constant-Last4}.

\begin{figure}[t]
    \centering
    \includegraphics[width=0.75\linewidth]{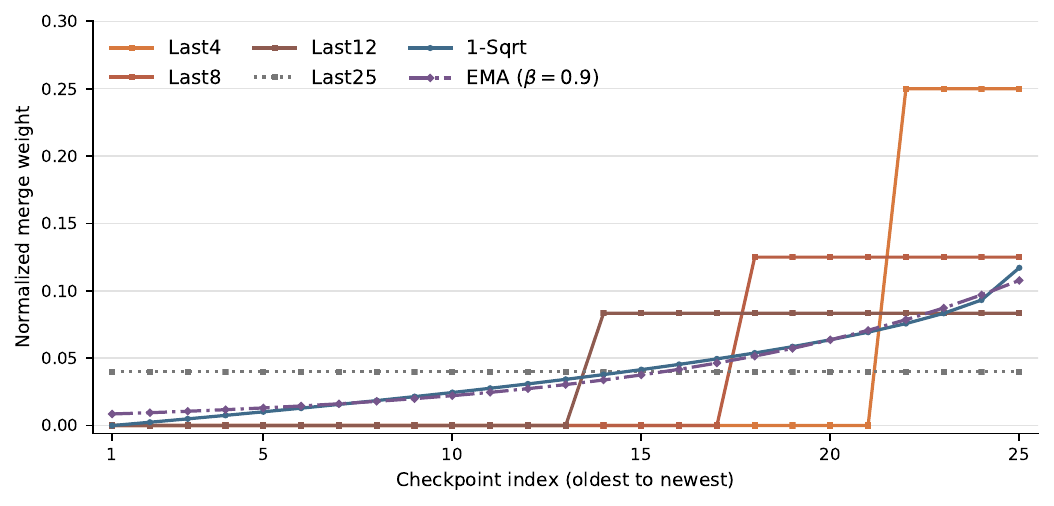}
    \caption{Normalized checkpoint-weight profiles on a common 25-position axis, including \textsc{Last4}, \textsc{Last8}, \textsc{Last12}, \textsc{Last25}, \textsc{1-Sqrt}, and \textsc{EMA} with $\beta=0.9$. Checkpoints are ordered from oldest to newest.}
    \label{fig:merge_weights}
\end{figure}

\paragraph{Model merging methods.}
For a weighted parameter average, we write
\begin{equation}
    \theta_{\mathrm{merge}}=\sum_{i=1}^{K}w_i\theta_i,
    \qquad
    w_i\geq 0,
    \qquad
    \sum_{i=1}^{K}w_i=1.
    \label{eq:merge_weighted_average}
\end{equation}
The representative deterministic weighting rules are illustrated in Figure~\ref{fig:merge_weights}.

\noindent\textbf{\textsc{Last$M$}.}
\textsc{Last$M$} assigns equal weight to the final $M$ consecutive checkpoints and zero weight to all earlier checkpoints:
\begin{equation}
    w_i=
    \begin{cases}
        1/M, & i>K-M,\\
        0, & i\leq K-M.
    \end{cases}
    \label{eq:merge_last_m}
\end{equation}
We evaluate several window sizes and use \textsc{Last4}, i.e., four weights of $0.25$, as the default method in the paper.
This rule performs a local trajectory average: it suppresses checkpoint-level fluctuations while avoiding the stale parameters of much earlier checkpoints.

\noindent\textbf{1-Sqrt.}
Let $t_i$ denote the training iteration of checkpoint $i$, and define its normalized age relative to the newest checkpoint as
\begin{equation}
    a_i=\frac{t_K-t_i}{t_K-t_1}.
    \label{eq:merge_normalized_age}
\end{equation}
The unnormalized and normalized \textsc{1-Sqrt} weights are
\begin{equation}
    \widetilde w_i=1-\sqrt{a_i},
    \qquad
    w_i=\frac{\widetilde w_i}{\sum_{j=1}^{K}\widetilde w_j}.
    \label{eq:merge_one_minus_sqrt}
\end{equation}
Thus, the oldest checkpoint receives zero weight, and the weight rises increasingly quickly near the end of the trajectory.
Unlike a rank-only rule, our implementation uses the actual iteration gaps through $t_i$.

\noindent\textbf{\textsc{EMA}.}
We implement post-hoc exponential moving averaging over all available checkpoints.
With one nominal checkpoint interval equal to $\Delta$ optimizer iterations, its weights are
\begin{equation}
    \widetilde w_i=\beta^{(t_K-t_i)/\Delta},
    \qquad
    w_i=\frac{\widetilde w_i}{\sum_{j=1}^{K}\widetilde w_j}.
    \label{eq:merge_ema}
\end{equation}
We sweep $\beta\in\{0.1,0.5,0.9\}$; Figure~\ref{fig:merge_weights} shows the representative $\beta=0.9$ setting used in the main comparison.
A smaller $\beta$ concentrates more mass on recent checkpoints, whereas a larger $\beta$ approaches a more uniform average over the trajectory.

Beyond weighted parameter averaging, we evaluate two representative
task-vector merging methods, \textsc{DARE} and \textsc{TIES}. For both
methods, we use the oldest selected checkpoint as the reference
$\theta_1$ and define the task vector for each source checkpoint as
$\delta_i=\theta_i-\theta_1$.

\noindent\textbf{\textsc{DARE}.}
We evaluate the \textsc{DARE-linear} variant~\citep{yu2024language}.
For each coordinate of each task vector, \textsc{DARE} samples an
independent Bernoulli mask with retention probability $\rho$, sets
dropped coordinates to zero, and rescales retained coordinates by
$1/\rho$. We then average the sparsified task vectors and add the
resulting update to $\theta_1$. We use random seed 42 and evaluate
$\rho\in\{0.1,0.5\}$.

\noindent\textbf{\textsc{TIES}.}
For each task vector and parameter tensor, \textsc{TIES} retains the
fraction $\rho$ of coordinates with the largest absolute magnitudes and
sets the remaining coordinates to zero. It then elects a sign for each
coordinate based on the sum of the trimmed task vectors, averages the
nonzero updates whose signs agree with the elected sign, and adds the
resulting update to $\theta_1$. We evaluate $\rho\in\{0.1,0.5\}$.

For visual clarity, Figure~\ref{fig:merge_methods} focuses on the better-
performing merging methods. We report \textsc{TIES} separately because
its best-performing configuration underperforms both \textsc{DARE} and
the final \textsc{Constant-Raw} checkpoint.

\paragraph{Interpreting the Advantage of Simple Averaging.}
TIES and DARE were developed primarily to control interference when combining task vectors from models with different fine-tuning objectives or acquired capabilities.
Our setting is structurally different: all checkpoints lie on one continuous training trajectory, so their updates are strongly correlated and temporally nested rather than representing independent capability directions.
In this regime, local or recency-weighted parameter averaging directly smooths trajectory noise while preserving the many small, consistently accumulated updates.
Magnitude trimming in \textsc{TIES} and random dropping in \textsc{DARE} can remove part of this shared trajectory signal, while their use of the full checkpoint range also assigns influence to relatively stale states.
Consistent with this interpretation, the best tested \textsc{TIES} setting ($\rho=0.5$) obtains a LongPPL of $3.7950$, which is worse than the raw final checkpoint at $3.6233$, while the best tested DARE setting ($\rho=0.5$) reaches $3.5217$ but remains behind \textsc{1-Sqrt} at $3.4724$ and \textsc{EMA} with $\beta=0.9$ at $3.4720$.

These results do not imply that \textsc{TIES} or \textsc{DARE} is generally inferior. Rather, they suggest that the mechanisms designed to resolve interference are less well suited to our setting, where checkpoints are highly correlated and lie along a single training trajectory. In this regime, simple weighted averaging is more effective, stable, and inexpensive.

\subsection{Sensitivity of Last4 to Checkpoint Spacing}
\label{app:merge-spacing}
The S5 \textsc{Constant-Last4} protocol in Section~\ref{sec:scaling_last4} averages the four most recent checkpoints with equal weight and uses a $1N$ gap between adjacent source checkpoints.
This choice balances evaluation stability against checkpoint storage, but it leaves open whether \textsc{Last4} LongPPL depends on that $1N$ spacing.

We therefore run a spacing ablation on the same S5 32K constant-learning-rate trajectory, with $D_s=450N$ and targets every $10N$ from $D_l=20N$ to $D_l=120N$.
For each target budget $T$ and spacing $s\in\{1N,2N,3N,4N,5N\}$, Last4($s$) uniformly averages the checkpoints $\{T-3s,\,T-2s,\,T-s,\,T\}$.
For example, the $1N$ window at $20N$ uses $17N/18N/19N/20N$, whereas the $5N$ window uses $5N/10N/15N/20N$.
As a control, we report the unmerged \textsc{Constant-Raw} checkpoint at the same target.
All protocols are evaluated on the same first $1{,}000$ 32K samples for LongPPL, ShortPPL, and PPL, so differences among spacings are not due to a changing evaluation set.

\begin{figure}[t]
    \centering
    \includegraphics[width=\linewidth]{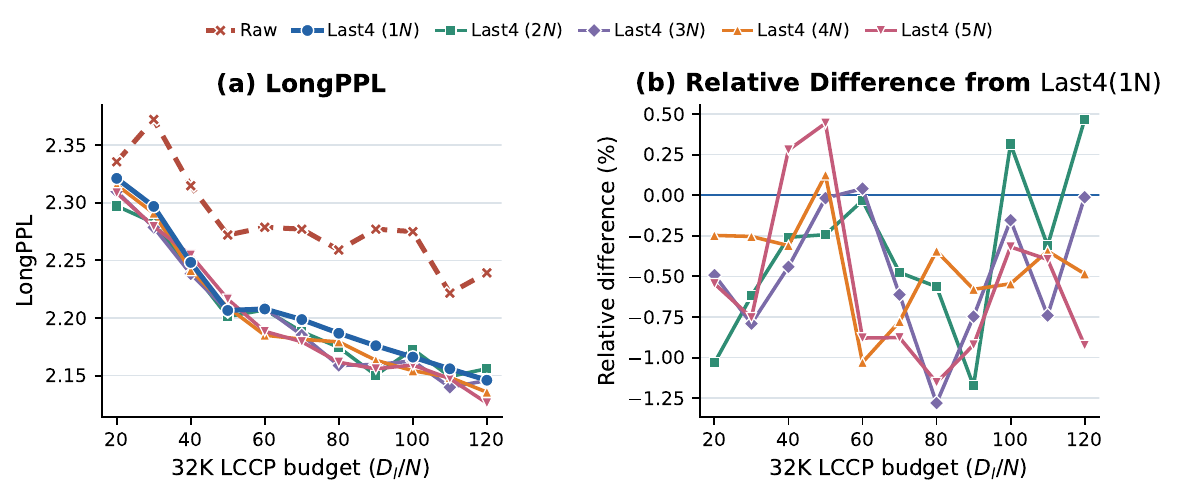}
    \caption{LAST4 window-spacing ablation on the S5 32K \textsc{Constant} trajectory. (a) LongPPL versus long-context budget. 
    (b) Relative difference from the default \textsc{Last4(1N)} window, computed as $100\times(\mathrm{LongPPL}-\mathrm{LongPPL}_{1N})/ \mathrm{LongPPL}_{1N}$ (\%).}
    \label{fig:merge_spacing}
\end{figure}

\begin{table}[t]
    \centering
    \setlength{\tabcolsep}{5pt}
    \caption{Mean LongPPL and ShortPPL over the eleven target budgets. $\Delta$ is the original-space difference against Last4(1N).}
    \label{tab:merge_spacing}
    \begin{tabular}{lrrrr}
        \toprule
        Protocol & \multicolumn{2}{c}{LongPPL} & \multicolumn{2}{c}{ShortPPL} \\
        \cmidrule(lr){2-3} \cmidrule(lr){4-5}
        & Mean & $\Delta$ vs $1N$ & Mean & $\Delta$ vs $1N$ \\
        \midrule
        Constant-Raw & 2.284 & $+0.074$ & 2.082 & $+0.041$ \\
        Last4 ($1N$) & 2.210 & $0.000$ & 2.041 & $0.000$ \\
        Last4 ($2N$) & 2.202 & $-0.008$ & 2.036 & $-0.005$ \\
        Last4 ($3N$) & 2.199 & $-0.011$ & 2.034 & $-0.007$ \\
        Last4 ($4N$) & 2.200 & $-0.010$ & 2.033 & $-0.008$ \\
        Last4 ($5N$) & 2.198 & $-0.012$ & 2.032 & $-0.009$ \\
        \bottomrule
    \end{tabular}
\end{table}

Figure~\ref{fig:merge_spacing}(a) and Table~\ref{tab:merge_spacing} give the same conclusion.
Merging improves over Raw at every one of the eleven budgets: mean LongPPL falls from $2.284$ to about $2.20$.
By contrast, stretching the source spacing from $1N$ to $5N$ changes the merged result only slightly.
The $2N$--$5N$ windows have mean LongPPL $0.008$ to $0.012$ below the default $1N$ window.
ShortPPL follows the same pattern: the improvement over Raw is about an order of magnitude larger than the differences among spacings.

Thus, the $1N$ Last4 protocol used for S5 is not a fragile choice. Within the tested $1N$--$5N$ range, Last4 is insensitive to checkpoint spacing, and the stabilization gain of merging over Raw dominates residual effects of the window geometry. Wider checkpoint spacing yields slightly lower perplexity on average, but the differences are small relative to the benefit of merging.

The ablation covers only this S5 trajectory and perplexity metrics, so we read it as a robustness check of the main-text $1N$ Last4 choice rather than as a universal guarantee for arbitrary spacings or downstream tasks.

\section{Downstream Evaluation Details}
\label{app:benchmark}
\subsection{RULER-32K Results by Subtask}
\label{app:ruler_32k}
\begin{figure*}[t]
    \centering
    \includegraphics[width=\textwidth]{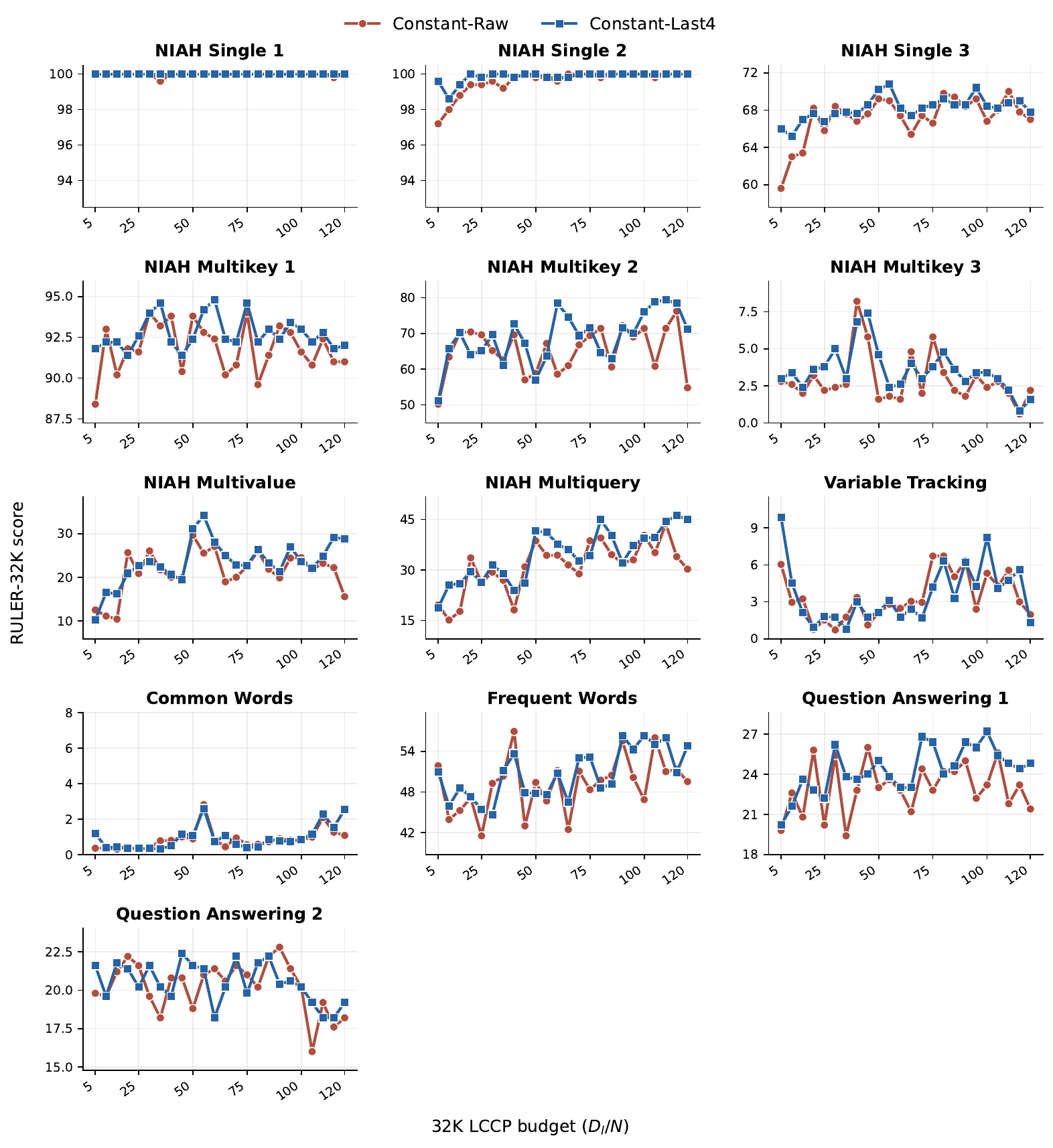}
    \caption{Per-subtask RULER-32K performance along the S5 32K LCCP trajectory. Scores are computed from 500 samples per subtask.}
    \label{fig:ruler32k_subtasks}
\end{figure*}

Section~\ref{sec:ruler32k} reports the aggregate RULER-32K score along the S5 32K LCCP trajectory. Figure~\ref{fig:ruler32k_subtasks} shows the corresponding scores on all 13 subtasks for both \textsc{Constant-Raw} and \textsc{Constant-Last4}.

\begin{figure*}[t]
    \centering
    \includegraphics[width=\textwidth]{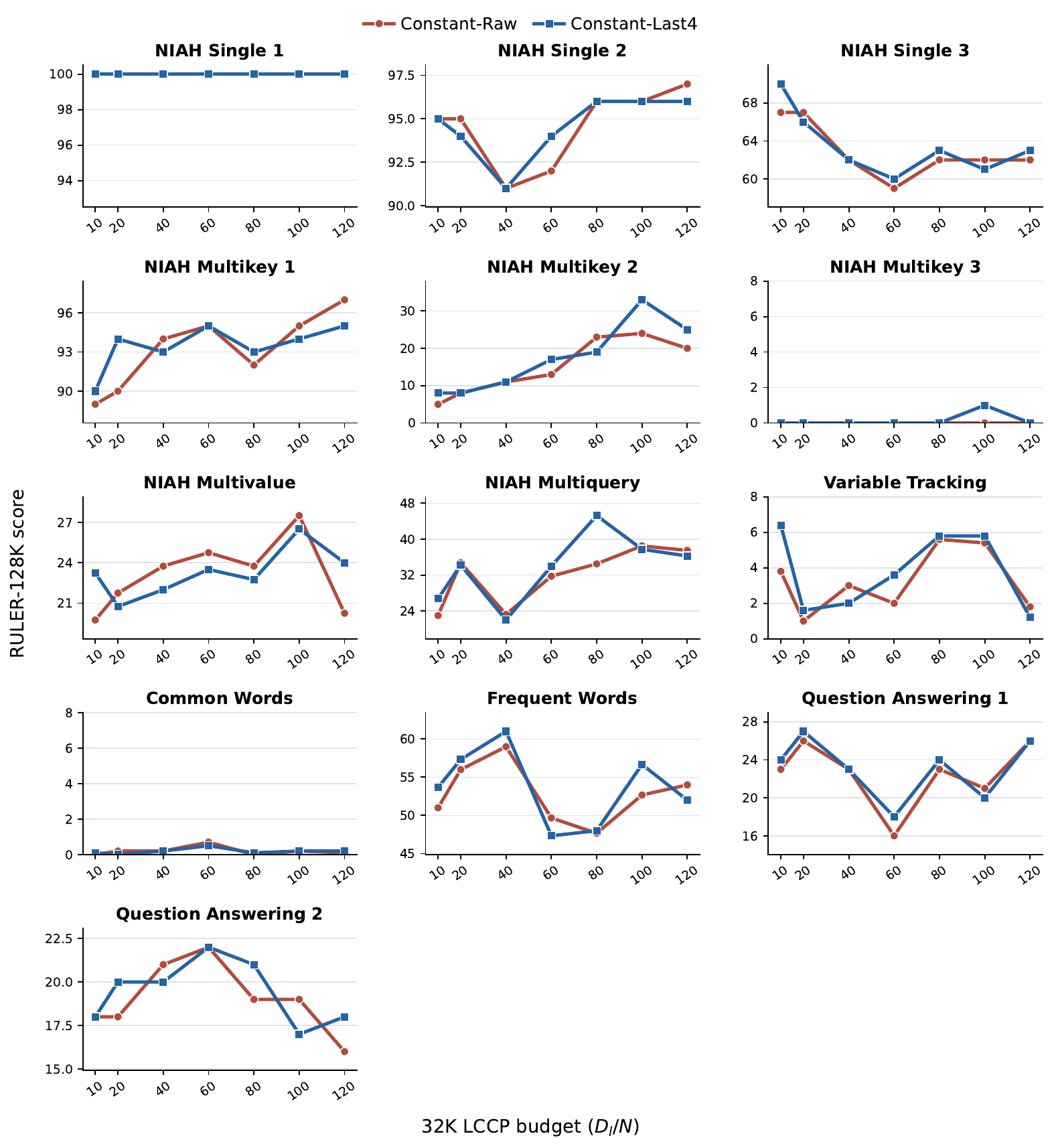}
    \caption{Per-subtask RULER-128K performance after the fixed 128K context-extension stage. The horizontal axis denotes the amount of 32K LCCP completed before context extension. All results use 100 fixed samples per subtask.}
    \label{fig:ruler128k_subtasks}
\end{figure*}

The easiest retrieval tasks saturate early: \texttt{niah\_single\_1} remains at ceiling throughout training, while \texttt{niah\_single\_2} and \texttt{niah\_multikey\_1} are already close to ceiling near the beginning of the trajectory. 
The clearest gains instead occur on intermediate-difficulty retrieval tasks. 
Under \textsc{Constant-Last4}, the scores of \texttt{niah\_multikey\_2}, \texttt{niah\_multivalue}, and \texttt{niah\_multiquery} increase from $51.20$, $10.20$, and $18.75$ at $5N$ to $71.20$, $28.75$, and $45.00$ at $120N$, respectively, despite checkpoint-level fluctuations. The corresponding \textsc{Constant-Raw} curves are less stable and show smaller endpoint gains.

The improvement is not uniform across all capabilities. The hardest multi-key retrieval setting, \texttt{niah\_multikey\_3}, and the more compositional Variable Tracking and Common Words tasks remain near the bottom of their respective scales. 
Scores on the QA tasks and Frequent Words also fluctuate rather than improve monotonically. Despite this uneven progress, gains on several intermediate-difficulty retrieval subtasks indicate continued benefits from additional LCCP.

\subsection{RULER-128K Results by Subtask}
\label{app:128k}
Similarly, Section~\ref{sec:128k} reports aggregate RULER-128K performance after applying the same 2.10B-token 128K context-extension stage to checkpoints selected from the S5 32K LCCP trajectory. We present the corresponding per-subtask scores in Figure~\ref{fig:ruler128k_subtasks}.

The 128K results exhibit a pattern similar to the 32K evaluation in the previous section. 
The score on \texttt{niah\_single\_1} remains at the ceiling of $100$, while scores on \texttt{niah\_single\_2} and \texttt{niah\_multikey\_1} remain high across all checkpoints.
The most visible transfer gains occur on \texttt{niah\_multikey\_2} and \texttt{niah\_multiquery}. 
As the preceding 32K LCCP budget increases from $10N$ to $120N$, \texttt{niah\_multikey\_2} rises from $5$ to $20$ for \textsc{Constant-Raw} and from $8$ to $25$ for \textsc{Constant-Last4}, reaching $33$ at the $100N$ Last4 checkpoint.
\texttt{niah\_multiquery} likewise rises from $23.00$ to $37.50$ for Raw and from $26.75$ to $36.25$ for Last4, with a Last4 peak of $45.25$ at $80N$.

Other tasks remain substantially harder at 128K. In particular, \texttt{niah\_multikey\_3}, Variable Tracking, and Common Words stay close to the floor, while Frequent Words and the two QA tasks vary across checkpoints without a reliable upward trend. 
The benefits of additional 32K LCCP therefore vary across subtasks and are not consistently monotonic, but gains in multi-key and multi-query retrieval contribute to the improvement in the aggregate 128K score.

\subsection{Evaluation after Long-Context and Agentic SFT}
\label{app:sft}

\begin{figure}[t]
\centering
\includegraphics[width=0.75\linewidth]{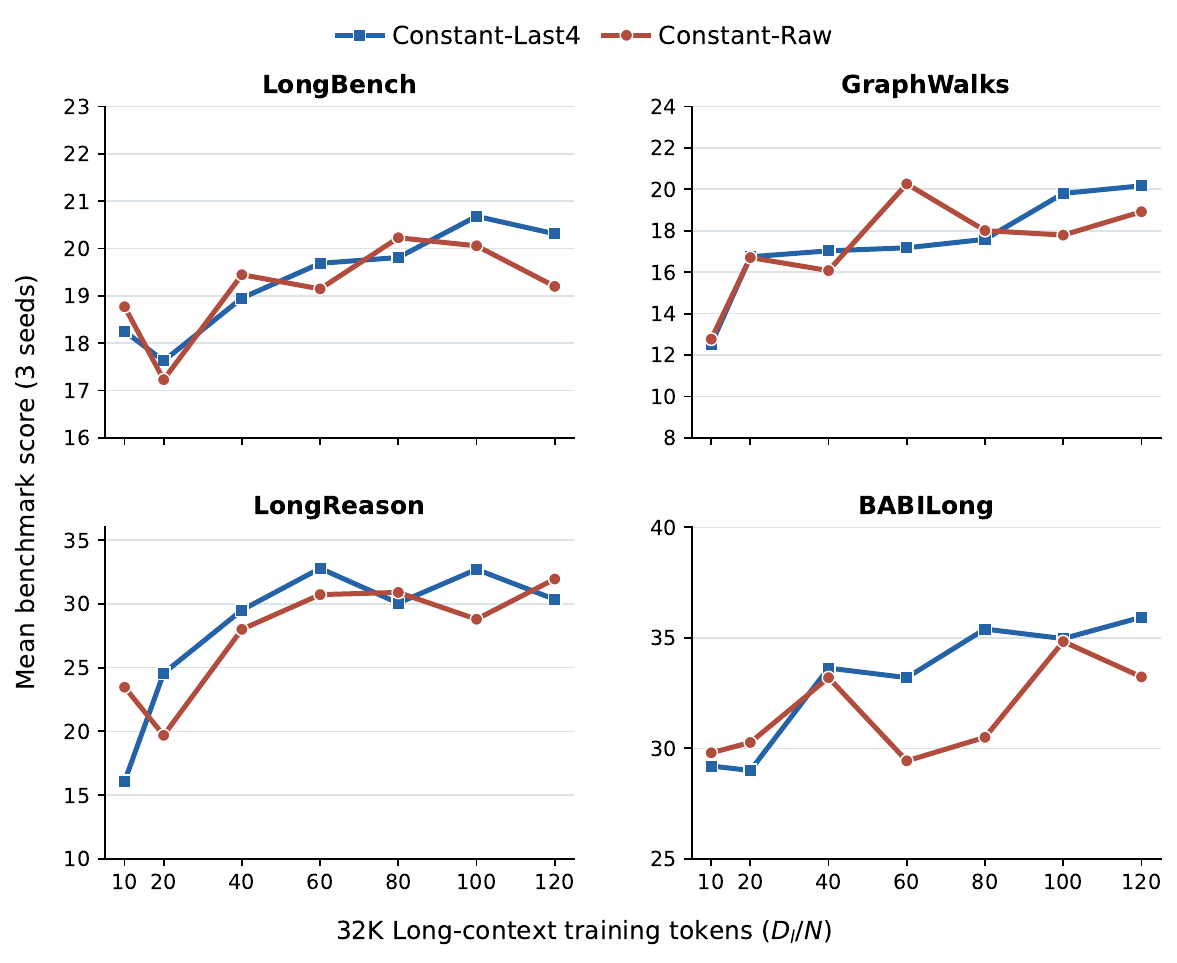}
\caption{Per-task performance after 2.10B tokens of SFT from checkpoints selected along the S5 32K LCCP trajectory for both \textsc{Constant-Last4} and \textsc{Constant-Raw}. Each point is the mean over three SFT runs trained with different data seeds.}
\label{fig:sft_subtask}
\end{figure}

\paragraph{Long-context SFT.}
The experiment in Section~\ref{sec:sft} applies a 2.10B-token SFT stage to checkpoints selected along the S5 32K LCCP trajectory.
For each starting checkpoint and checkpoint-selection method, we train three SFT models using different data seeds. We then evaluate each model on LongBench, GraphWalks, LongReason, and BABILong. Figure~\ref{fig:sft_subtask} reports the mean score over the three seeds for each benchmark separately, complementing the equal-weight four-task averages reported in the main text.

To test whether this benefit persists with a larger SFT budget, we extend SFT to 4.20B tokens (4,000 steps) from the same \textsc{Constant-Last4} and \textsc{Constant-Raw} starting checkpoints. Apart from doubling the SFT duration, we keep the training and evaluation protocols consistent with the 2.10B-token experiment. We evaluate LongBench, GraphWalks, LongReason, and BABILong and report both the individual scores and their equal-weight average. Due to the additional training cost, these 4,000-step experiments use only one data seed and should therefore be interpreted as supporting evidence rather than a seed-robust comparison.

\begin{figure*}[t]
\centering
\includegraphics[width=0.75\textwidth]{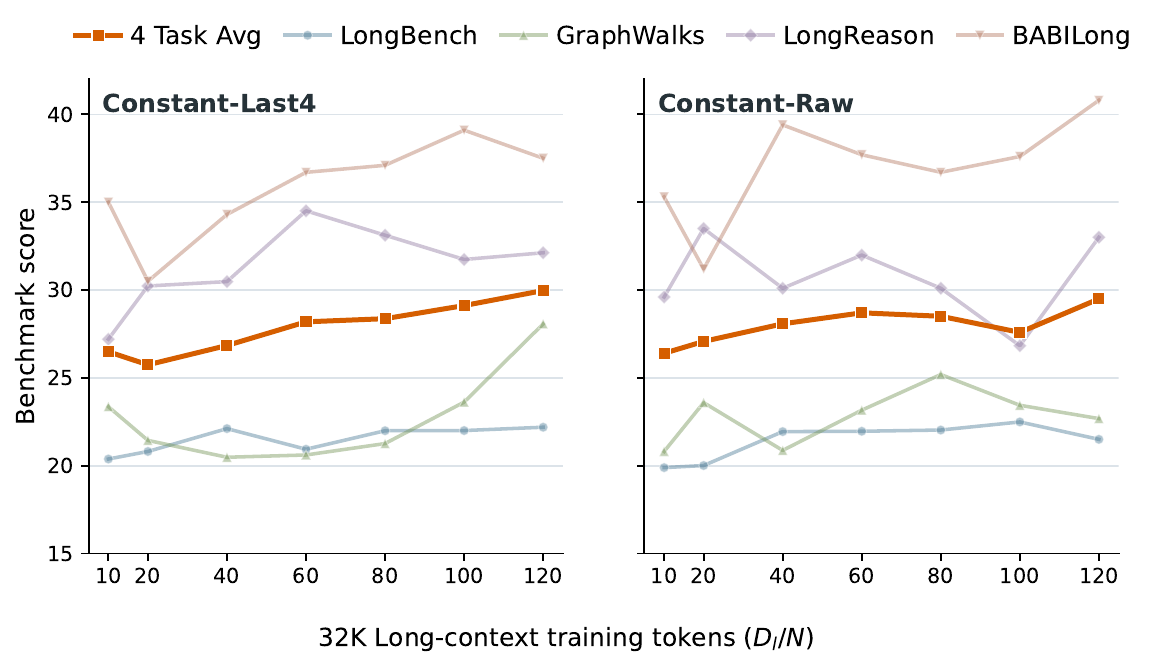}
\caption{Performance after 4.20B-token SFT from S5 checkpoints at varying 32K LCCP budgets. Individual benchmark scores and their equal-weight average follow the 2.10B-token evaluation protocol. These single-data-seed results serve as a supplementary robustness check.}
\label{fig:sft_4k}
\end{figure*}

\begin{figure}[htbp]
\centering
\includegraphics[width=0.75\linewidth]{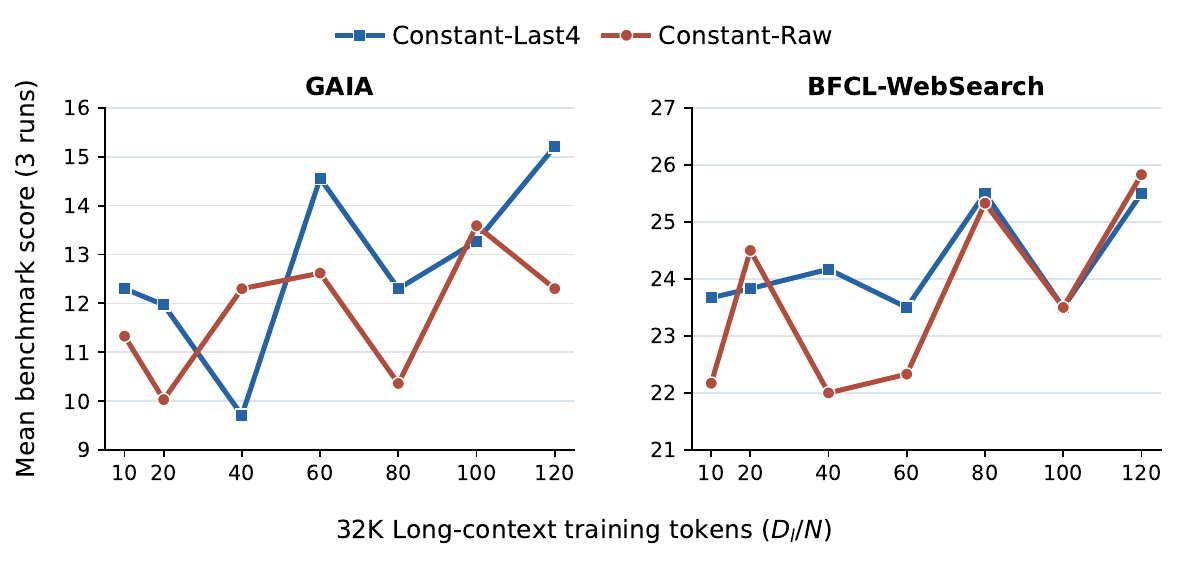}
\caption{GAIA and BFCL-WebSearch performance after 4.20B-token agentic SFT along the S5 32K LCCP trajectory. Each point averages three evaluation runs.}
\label{fig:search_agent_subtasks}
\end{figure}

Figure~\ref{fig:sft_4k} provides preliminary evidence that the main conclusion persists after doubling the SFT budget. For \textsc{Constant-Last4}, the four-benchmark average initially decreases from $26.49$ at $10N$ to $25.74$ at $20N$, but then rises consistently to $29.97$ at $120N$. For \textsc{Constant-Raw}, the average fluctuates more across intermediate checkpoints but still increases from $26.40$ to $29.49$ over the same range. Individual benchmarks remain noisy, so the gain is not uniformly monotonic on every task. Nevertheless, the overall trend is consistent with the hypothesis that the benefits of additional LCCP persist when the SFT budget is doubled, while the single-seed setting limits the strength of this conclusion.

\paragraph{Agentic SFT.}

The agentic SFT experiment in Section~\ref{sec:sft} applies a 4.20B-token SFT stage to checkpoints selected along the S5 32K LCCP trajectory. 
We evaluate each model on three long-horizon agentic benchmarks, covering two task types: search agent (GAIA and the web search category of BFCL~v4(hereafter, BFCL-WebSearch)) and multi-turn tool calling ($\tau^2$-telecom). In all three, the context accumulates over interactions and reaches tens of thousands of tokens (Table~\ref{tab:agent_context_len}), exercising long-context capability. Inference uses the thinking response format with temperature 0.6 and a maximum context length of 32,768 tokens, and we report the average score over three independent runs (avg@3).

For GAIA, we use the text-only subset of the validation set, totaling 103 examples, and evaluate with the MiroThinker agent framework~\citep{team2025mirothinker}. The agent runs a tool-calling loop of up to 200 turns with web search, web-page reading, and Python-interpreter tools. To fit long trajectories into the context budget, only the five most recent tool results are kept in the prompt while earlier ones are replaced by a placeholder. Final answers are judged by GPT-4.1 against the reference.

For BFCL-WebSearch, the benchmark consists of 100 multi-hop questions whose answers must be retrieved from the live web, and we follow the official protocol for evaluation. The agent runs a native function-calling loop of up to 20 steps with a web search tool and a web-page reading tool. We report the mean accuracy over its two settings (\textit{base}, where search results include text snippets, and \textit{no-snippet}, where the model must open web pages instead).

For $\tau^2$-telecom, we evaluate on the telecom domain of $\tau^2$-bench, totaling 114 tasks. The user is simulated by Qwen3-235B-A22B-Instruct-2507~\citep{qwen3technicalreport} with temperature 0, and each dialogue is capped at 100 steps. A trial scores 1 only if all task-specified checks on the final environment state and required tool calls pass, and 0 otherwise.

Table~\ref{tab:agent_context_len} summarizes the context lengths observed during evaluation. For GAIA, we report both the final context (the input of the last LLM call) and the peak context (the largest input over the course). BFCL-WebSearch and $\tau^2$-telecom keep the full history in context, so the final context equals the peak. BFCL trajectories exceeding the context limit are scored 0 and excluded from the table.

\begin{table}[h]
\centering
\caption{Context lengths (tokens) of agentic evaluation trajectories.}
\label{tab:agent_context_len}
\begin{tabular}{llrrrrr}
\toprule
Benchmark & Context & Avg & P50 & P75 & P90 & Max \\
\midrule
GAIA         & final        &  7,036 &  6,603 & 10,470 & 13,361 & 31,512 \\
GAIA         & peak         & 12,533 & 12,749 & 21,074 & 21,167 & 31,512 \\
BFCL-WebSearch & final (peak) & 16,259 & 15,615 & 22,977 & 28,900 & 32,758 \\
$\tau^2$-telecom & final (peak) & 10,844 & 10,543 & 11,376 & 12,376 & 32,766 \\
\bottomrule
\end{tabular}
\end{table}

\section{Additional Analyses}
\label{app:additional_analysis}
\subsection{Comparing LongPPL and ShortPPL Improvement Rates}
\label{app:longppl_analysis}

Although LongPPL is widely used as a measure of long-context performance~\citep{willette2026delta,song2026towards} and correlates well with a range of long-context downstream tasks~\citep{qiao2026rethinking,fang2025wrong}, it is still a perplexity-based metric. A lower LongPPL need not result exclusively from better use of distant context: it can also reflect broad next-token prediction gains, adaptation to the evaluation domain, or changes in local token statistics. We therefore compare changes in LongPPL and ShortPPL to examine whether tokens with different dependence on distant context improve at different rates during LCCP.

We use ShortPPL as a matched control metric. It is computed on tokens whose prediction has limited dependence on distant context, and thus provides a proxy for perplexity changes that do not require long-range information. ShortPPL is not a pure measure of general language-modeling ability, but it helps capture broad improvements under the same model, data, and evaluation procedure. Along the S5 \textsc{Constant-Last4} trajectory, we fit the same saturating power-law form independently to both metrics over 24 checkpoints, from $D_l=5N$ to $D_l=120N$ (approximately $100$B tokens):
\[
\log M(D_l)=E_M+A_M D_l^{-\alpha_M}, \qquad M\in\{\mathrm{LongPPL},\mathrm{ShortPPL}\}.
\]

For metric $M$, we define the log-scale elasticity
\[
e_M(D_l)=\left|\frac{\partial\log M}{\partial\log D_l}\right|.
\]
For the fitted form, this is $e_M(D_l)=\alpha_M A_M D_l^{-\alpha_M}$, so it measures the relative rate of improvement at a given training budget. We then compare the two metrics using
\[
R(D_l)=\frac{e_{\mathrm{Long}}(D_l)}{e_{\mathrm{Short}}(D_l)}.
\]
This ratio is a descriptive comparison: it indicates whether LongPPL is changing faster than the matched ShortPPL control, but it is not an exact causal decomposition of the total perplexity gain.

\begin{figure}
    \centering
    \includegraphics[width=0.75\linewidth]{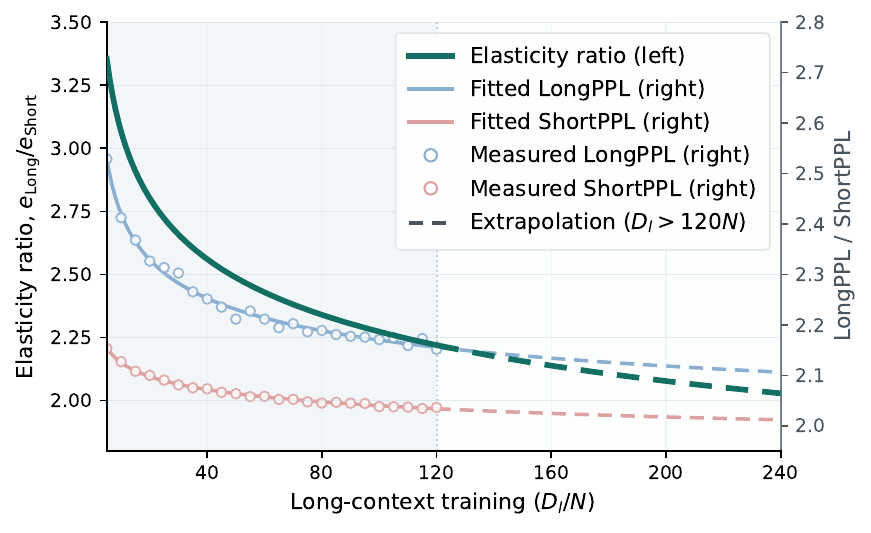}
    \caption{Ratio of the log-scale elasticities of LongPPL and ShortPPL along the S5 \textsc{Constant-Last4} LCCP trajectory. Markers show measurements through ($D_l=120N$). Solid curves show fits over the measured range; dashed curves extend these fits to ($D_l = 240N$).}
    \label{fig:longppl_slope_ratio}
\end{figure}

As shown in Figure~\ref{fig:longppl_slope_ratio}, $R(D_l)$ is largest at the beginning of the trajectory and gradually decreases as LCCP proceeds. Numerically, it declines from approximately $3.35$ at $5N$ to $2.22$ at $120N$, but remains above $2$ throughout the observed range. Thus, LongPPL improves substantially faster than ShortPPL, although the difference in the relative improvement rates of LongPPL and ShortPPL narrows as training proceeds.
In conjunction with the downstream evaluations in Section~\ref{sec:benchmark}, the result supports using LongPPL as a useful proxy for long-context performance during extended LCCP.

\subsection{Effects of Checkpoint Merging across Training Stages}
\label{app:merge_analysis}

\begin{table*}[t]
    \centering
    \caption{Mean performance of \textsc{Constant-Raw} and
    \textsc{Constant-Last4} over comparable evaluation points. The training stage identifies where each metric is evaluated. For all rows, $\Delta=\textsc{Last4}-\textsc{Raw}$; negative values indicate improvement for LongPPL, whereas positive values indicate improvement for other task scores.}
    \label{tab:raw_last4_summary}
    \small
    \setlength{\tabcolsep}{6pt}
    \begin{tabular}{llrrrr}
        \toprule
        Metric & Training stage & $n$ & \textsc{Raw} & \textsc{Last4} & $\Delta$ \\
        \midrule
        32K LongPPL & 32K LCCP & 24 & 2.312 & 2.241 & $-0.071$ \\
        RULER-32K & 32K LCCP & 24 & 44.29 & 45.48 & $+1.19$ \\
        RULER-128K & 128K LCCP & 7 & 39.88 & 40.48 & $+0.60$ \\
        SFT-4-avg (2.10B) & SFT & 7 & 23.91 & 24.42 & $+0.51$ \\
        SFT-4-avg (4.20B) & SFT & 7 & 27.98 & 27.82 & $-0.16$ \\
        Agentic-3-avg (4.20B) & SFT & 7 & 35.21 & 35.37 & $+0.15$ \\
        \bottomrule
    \end{tabular}
\end{table*}

During extended LCCP, \textsc{Constant-Last4} lowers LongPPL and improves average RULER-32K performance relative to \textsc{Constant-Raw}.
Table~\ref{tab:raw_last4_summary} summarizes these differences and compares performance after subsequent training initialized from the corresponding raw and merged checkpoints.
After 128K context extension, models initialized from \textsc{Constant-Last4} outperform those initialized from \textsc{Constant-Raw} by $0.60$ points on average on RULER-128K.
Similarly, after 2.10B tokens of SFT, the four-benchmark average is $0.51$ points higher for \textsc{Constant-Last4}. 
However, with a larger SFT budget of 4.20B tokens, this difference narrows and slightly favors \textsc{Constant-Raw} by $0.16$ points. In agentic SFT, \textsc{Constant-Last4} retains a small $0.15$-point advantage over \textsc{Constant-Raw}. Thus, the advantage of checkpoint merging observed during LCCP does not consistently persist through subsequent training.

A related finding is that learning-rate decay can improve pre-training metrics while yielding worse performance after SFT than training without decay~\citep{yano2026pre}.
Although this concerns learning-rate schedules rather than checkpoint merging, it similarly highlights that better performance before SFT need not imply better performance afterward.
Our limited experiments do not establish when merging advantages persist.
One tentative hypothesis is that these advantages may survive a modest amount of additional training or with a similar objective, as in context extension, but diminish or reverse with longer training or larger shifts in training data and objectives, as in SFT.
Further controlled experiments are needed to distinguish these factors.

We therefore primarily view checkpoint merging as a stable \emph{observation} mechanism for assessing long-context performance along an extensible constant-learning-rate trajectory, rather than as a strategy for consistently improving subsequent training outcomes.
In this protocol, merging is used for evaluation, while continued LCCP proceeds from the latest raw checkpoint.

\subsection{Dynamics of the Retrieval-Head during LCCP}
\label{app:retrieval_head_metrics}

Previous work suggests that retrieval-head metrics can serve as mechanistic probes of model development over the course of LCCP~\citep{liang2026revealing}. 
Following the definition in that work, we measure the copy-retrieval behavior of individual attention heads. For each teacher-forced answer token, a head records a retrieval hit when its maximum-attended input position lies within the reference sentence and contains the same token\_id as the current target token. The per-instance score of a head is the fraction of answer tokens producing such hits, and its checkpoint-level retrieval score is obtained by averaging the per-instance scores over our 2,400 LongBio single-needle instances~\citep{yang2026controllable}. A head is classified as a retrieval head if its checkpoint-level score exceeds $0.1$.
For longitudinal comparison, we identify 10 retrieval heads at the raw $5N$ checkpoint and track the mean score of this fixed set at all subsequent checkpoints. The retrieval-head count, by contrast, is recomputed independently at each checkpoint as the number of heads whose checkpoint-level score exceeds $0.1$.
\begin{figure}[t]
    \centering
    \includegraphics[width=1.0\linewidth]{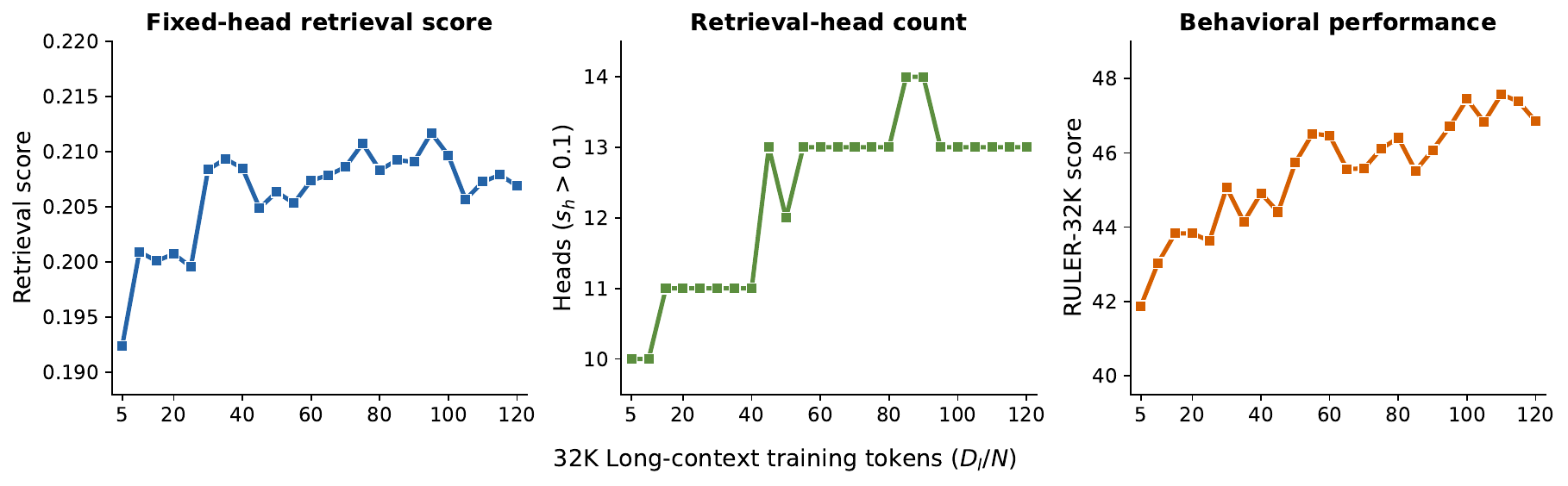}
    \caption{Retrieval-head metrics and RULER-32K performance along the S5
    \textsc{Constant-Last4} trajectory. The three square panels show the
    fixed-head retrieval score, retrieval-head count, and RULER-32K score.}
    \label{fig:retrieval_head_monitoring}
\end{figure}

As shown in Figure~\ref{fig:retrieval_head_monitoring}, the retrieval-head metrics capture substantial learning during the early stages of training. The mean score of the fixed retrieval heads rises rapidly, while the number of heads above the threshold increases from 10 to 13--14. After around $D_l=30N$, however, the fixed-head score reaches a noisy plateau and exhibits no sustained upward trend.
This plateau does not coincide with saturation in behavioral performance. The RULER-32K score continues to improve up to $D_l=120N$, with additional gains on several multi-key and multi-query tasks. 
These results suggest that the selected retrieval-head probes do not capture all behavioral improvements that occur later in LCCP.
Retrieval-head metrics are therefore useful as early-stage diagnostic signals, but they should be considered alongside LongPPL and behavioral evaluations rather than used alone to determine whether LCCP training has saturated.

\end{document}